\documentstyle[11pt,twoside,chicago,tree-dvips]{article}

\input{psfig}

\newcommand{\mcc}[2]{\multicolumn{#1}{c}{#2}}

\newcommand{\n}[2]{\node{#1}{#2}}
\newcommand{\nc}[2]{\nodeconnect[b]{#1}[t]{#2}}

\def\lf{\vspace{1ex}} 
\newcommand{\hs}[0]{\hspace{1em}}

\newcommand{\avmsz}{\scriptsize}

\newlength{\typeindent}
\newcommand{\avmt}[2]{\renewcommand{\arraystretch}{1.2}
\sbox{\boxtmp}{%
\settowidth{\typeindent}{\avmsz\it{#1}}\hspace{\typeindent}%
$\left[\mbox{\avmsz\begin{tabular}{@{\rm}l@{\,\hspace{0.8em}}l@{}}#2\\[-3ex]%
\makebox[0pt][r]{\raisebox{-2ex}[0pt][0pt]{\avmsz\it #1\
\,}}\end{tabular}}\right]$}
\begin{tabular}{@{}l@{}}
\usebox{\boxtmp} \\ \\[-2ex]
\end{tabular}\renewcommand{\arraystretch}{1}}

\newcommand{\avm}[1]{\renewcommand{\arraystretch}{1.2}\mbox{%
$\left[\mbox{\avmsz\begin{tabular}{@{\rm}l@{\,\hspace{0.8em}}l@{}}#1
\end{tabular}}\right]$}\renewcommand{\arraystretch}{1}}

\newcommand{\avmNo}[1]{\renewcommand{\arraystretch}{1.2}\mbox{%
$\left[\mbox{\avmsz\begin{tabular}{l}\rm#1
\end{tabular}}\right]$}\renewcommand{\arraystretch}{1}}

\newcommand{\myavm}[1]{\renewcommand{\arraystretch}{1.2}
\mbox{\(\left[\avmsz\begin{tabular}{@{\rm}l@{\,\hspace{0.8em}}l@{}}%
#1\end{tabular}\right]\)}\renewcommand{\arraystretch}{1}}

\newcommand{\idx}[1]{\mbox{\fbox{\tiny #1}\hspace{0.1em}}}

\newsavebox{\boxtmp}

\newcounter{myfignr}

\newcommand{\myfigPB}[3]{\myfigRef{#1}\begin{center} #3 \end{center} \myfigTitle{#2}}

\newcommand{\myfig}[3]{\myfigRef{#1}
\parbox[t]{\textwidth}{\begin{center} #3 \end{center} \myfigTitle{#2}}}

\newcommand{\myfigRef}[1]{\refstepcounter{myfignr}\label{#1}
  \vspace*{0.4ex}} 

\newcommand{\myfigTitle}[1]{\begin{center}
  \par\vspace*{0.2ex} \rm \normalsize Figure \arabic{myfignr}:
  #1\end{center} \vspace*{0.6ex}}

\newcommand{\mysection}[1]{\lf {\bf #1 \nopagebreak \par \nopagebreak}}
\newcommand{\mysubsection}[1]{\lf {#1 \nopagebreak \\ \nopagebreak}}
\newcounter{hilfszaehler}

\newcommand{\nd}[2]{\mbox{ \( \{\!_{\rm #1} \;\; #2 \} \) }}
\newcommand{\dis}[0]{\mbox{\(\; ; \;\)}}

\newcommand{\tassign}{\mbox{\(\sim\)}}
\newcommand{\gleich}{\mbox{\(=\)}}
\newcommand{\ungleich}{\mbox{\(\neq\)}}
\newcommand{\und}{\mbox{\(\wedge\)}}
\newcommand{\oder}{\mbox{\(\vee\)}}
\newcommand{\nicht}{\mbox{\(\neg\)}}
\newcommand{\impl}{\mbox{\(\rightarrow\)}}

\newcommand{\nfin}{\typ{n\_fin}} \newcommand{\fin}{\typ{fin}}
\newcommand{\finv}{\typ{fin\_v}} \newcommand{\nfinv}{\typ{n\_fin\_v}}
\newcommand{\plus}{\typ{plus}} \newcommand{\minus}{\typ{minus}}
\newcommand{\bool}{\typ{bool}}

\newcommand{\psimpl}{\mbox{\(\rightarrow\)}}

\newcommand{\idimpl}{\psimpl}

\newcommand{\interpret}[0]{\parbox[t]{5em}{\centering \(\longrightarrow\)
    \\[-1.4ex] {\scriptsize \it Interpretation\/}}}

\newcommand{\oneOne}[0]{\parbox[t]{7em}{\centering \(\longrightarrow\)
    \\[-1.4ex] {\scriptsize \it \parbox[t]{7em}{\centering One-to-one
        \\[-1ex] Correspondence}\/}}}

\newcommand{\lst}[2]{\(\left<\left. \mbox{#1} \hspace{0.4em} \right|
  \hspace{0.34em} \mbox{#2} \; \right>\)}

\newcommand{\lstI}[1]{\(\left< \mbox{#1} \; \right>\)}
\newcommand{\elst}[0]{\(\left<\right>\)}

\newcommand{\Xlst}[2]{\(<\mbox{#1} \hspace{0.4em}|
  \hspace{0.34em} \mbox{#2} >\)}

\newcommand{\XlstI}[1]{\(< \mbox{#1} >\)}
\newcommand{\Xelst}[0]{\(<>\)}
\def\|{\mbox{\(|\)}}
\def\typ#1{\mbox{{\it #1}}}
\def\hpsgII{\mbox{HPSG \hspace*{-1.3ex} II}}
\newcommand{\path}[1]{\mbox{\({\scriptsize \leadsto}\!\)#1}}

\begin{document}


\title{\vspace*{-8ex}On Implementing an HPSG theory\thanks{I am grateful to Thilo
    G\"otz, Erhard Hinrichs, Tilman H\"ohle, and Paul King for
    valuable discussion and advice, and to Tsuneko Nakazawa for many
    interesting discussions of concepts in and alternatives to the
    theory proposed by Erhard Hinrichs and herself. For comments on a
    previous version of the paper I want to thank John Griffith, Tibor
    Kiss, Esther K\"onig-Baumer, Frank Morawietz, J\"urgen Pafel,
    Frank Richter, and Susanne Riehemann.}\\[1ex]\large Aspects of the
  logical architecture, the formalization, and the implementation\\ of
  head-driven phrase structure grammars\vspace*{2ex}}

\author{Walt Detmar Meurers\\ \small Sonderforschungsbereich
  340, Universit\"at T\"ubingen\\[-3pt] \small Kleine Wilhelmstr.\ 
  113, 72074 T\"ubingen, Germany \\[-3pt]\small email:
  dm@sfs.nphil.uni-tuebingen.de\vspace*{1ex}}

\date{{\small In: Erhard W.\ Hinrichs, W.\ Detmar Meurers, and Tsuneko
    Nakazawa:\\ {\it Partial-VP and Split-NP Topicalization in German --
      An HPSG Analysis and its Implementation}.\\  Arbeitspapiere des
    {SFB} 340 Nr.\ 58, Universit\"at T{\"u}bingen}, 1994}
 
\maketitle


\begin{center}
{\large \bf Table of Contents}
\end{center}

\contentsline {section}
{\numberline {\ref{intro-section}}Introduction}
{\pageref{intro-section}}

\contentsline {section}
{\numberline {\ref{ontology-section}}The HPSG ontology: A sketch of the logical foundations}
{\pageref{ontology-section}}

\contentsline {section}
{\numberline {\ref{architecture-section}}The HPSG architecture: A linguist's view on the feature logic toolkit}
{\pageref{architecture-section}}

\contentsline {subsection}
{\numberline {\ref{signature-section}}Declaring the domain of a grammar}
{\pageref{signature-section}}

\contentsline {subsection}
{\numberline {\ref{constraint-section}}Constraining the domain}
{\pageref{constraint-section}}

\contentsline {section}
{\numberline {\ref{ling-constraint-section}}Modules of an HPSG theory: Special linguistic types of constraints and\\choices for encoding them}
{\pageref{ling-constraint-section}}

\contentsline {subsection}
{\numberline {\ref{lexicon}}Specifying the lexicon}
{\pageref{lexicon}}

\contentsline {subsection}
{\numberline {\ref{struc-licensing}}Structure licensing}
{\pageref{struc-licensing}}

\contentsline {subsection}
{\numberline {\ref{principles}}Expressing the principles}
{\pageref{principles}}

\contentsline {section}
{\numberline {\ref{specific-theory}}Relating the specific linguistic theory to its implementation}
{\pageref{specific-theory}}

\contentsline {section}
{References}
{\pageref{references}}

\contentsline {section}
{Appendix A: The grammar}
{i}

\contentsline {section}
{Appendix B: A test run}
{xlvii}

\newpage

\section{Introduction}
\label{intro-section}

The paper documents the implementation of an HPSG theory for German in
a constraint resolution system. It illustrates the reasoning and the
choices involved in the formalization and implementation of HPSG
grammars.  As basis for the reasoning, a discussion of the logical
setup of the HPSG architecture as proposed in \citeN{hpsg2}
(henceforth \hpsgII) is supplied. The relationship between a
linguistic theory and its implementation is explored from a linguistic
point of view.

The grammar implemented covers the phenomena of aux-flip and partial
verb phrase topicalization in the three sentence types of German: verb
first, verb second, and verb last. It closely follows the analyses
proposed in \citeN{Hinrichs&Nakazawa91} and
\citeN{Hinrichs&Nakazawa94} (henceforth HN), except for a few cases
noted in section~\ref{specific-theory}.

The implementation is based on two systems: Bob Carpenter and Gerald
Penn's ALE, and the Troll system developed by Dale Gerdemann and Thilo
G\"otz based on a logic by Paul King. The discussion of computational
architectures in this paper also includes references to other systems,
such as Martin Emele and R\'emi Zajac's TFS, or the CUF system
developed by Jochen D\"orre, Michael Dorna, and Andreas
Eisele.\footnote{For a brief overview of the main characteristics of
  TFS and CUF including some comments on the implications for the work
  of the linguist cf.~\citeN{Meurers&Goetz93}.
  \citeN{Manandhar93} is a comparison of ALE, CUF, and TFS regarding
  the expressive power of the type system, the definite clauses, and
  the control scheme.}

The structure of the paper is as follows: The remaining part of this
introduction briefly reports on the why and how of the project: the
motivations behind the implementation, and the systems used. The
second section describes the logical setup of HPSG: the description
language and its interpretation. The third section deals with the
feature logic building blocks of the HPSG architecture: the
declaration of the domain of the grammar and the possibilities for
formulating the constraints making up the linguistic theory. The focus
is on the implications of the logical setup for the work of the
linguist and examples from the implemented grammar are supplied.  The
fourth section goes through the linguistic modules of an HPSG theory:
the lexicon, the specification of constituent structure, and the
grammatical principles.  The choices involved in formalizing them for
implementation are discussed and illustrated with examples. The fifth
section discusses the differences between the linguistic theory in HN
and the implemented grammar and gives some technical comments on the
implemented grammar. The complete grammar including a collection of
test sentences and a test run are included in the appendix.

\subsection{Motivation}

There are at least two different motivations for implementing HPSG theories.

From a computational point of view a complex theory serves well as a
test for the feature logic systems being developed. It can illustrate
performance and theoretic devices, such as named disjunction or full
negation, or more specific mechanisms, like definite clause
attachments on lexical entries. The line of thinking behind such
implementations starts from the devices, explores their usefulness in
compactly encoding linguistic generalizations, and then illustrates
the computational setup in an implementation. Even though the
intentions under this approach are oriented towards computation, the
work can provide feedback for linguistics in supplying new theoretical
devices for expressing linguistic theories.

The intention of a linguist on the other hand is to use the
development of an implementation as a tool to provide feedback for a
rigid and complete formalization of a linguistic theory.  Under this
approach the work starts out from the theory and tries to put the
mechanisms proposed in it to work together in a grammar in an elegant
and correct fashion.

Since the work presented here started out from a linguistic theory, it
is closer to the second view. Nonetheless, the implementation also
shows that both systems employed, ALE and Troll, can be used to
express and process rather complex grammars. One of the effects of the
linguistic approach pursued is that the implemented grammar tries to
stay close to the notation and formulation used in the linguistic
theory. The computational aspect causes some limitations here, the
major one being the departure from ID/LP format due to the phrase
structure based systems used.

\subsection{The systems used}

This section briefly introduces the computational setup of the
implementation. A detailed discussion of the theoretical properties
and differences between the two systems (which made it interesting to
try to use the same grammar in both approaches) is included in the
next sections, where the HPSG architecture and its logical foundation
are discussed.

The purpose of the Troll system is to efficiently implement a logic
that gives a denotational semantics to typed feature structures as
normal form descriptions as opposed to thinking of feature structures
as models\footnote{This ontological difference between the logic of
  \citeN{King89} and that of \citeN{Carpenter92} is elaborated in
  section~\ref{ontology-section}.}. This allows unfilling of feature
structures since there is a clear notion of semantic equivalence of
different feature structures (cf.~\citeN{Goetz94}).  Unfilling is
an operation which removes arcs not constraining the denotation of a
feature structure.  Technically it allows for efficient representation
of descriptions, while for the linguist it serves to suppress the
information that can be inferred from the signature.  Troll also
assumes a closed world interpretation of type hierarchies which seems
natural for linguistics and HPSG in particular.

Due to the nature of Troll, the notation only represents the concepts
relevant from the viewpoint of logic. For the implementation to
reflect the linguistic theory, several extensions are necessary:
Macros\footnote{The use of macros in the lexicon and the problems
  involved in replacing them with an appropriate formal mechanism are
  discussed in section~\ref{lex-classes}.} serve to follow linguistic
notation and avoid unnecessary, unclear and error prone repetition.
Lexical rules, even though their theoretical status is not completely
clear, are used in many HPSG publications including the one
implemented, and therefore need to be part of the language.  Finally,
to enable debugging of a non-trivial grammar, a source level debugger
allowing very selective output of information as well as good
displaying and editing possibilities for the extensive feature
structures is necessary.

Since ALE\footnote{For a description of the ALE system, the reader is
  referred to \citeN{ALEmanual}. ALE Version 1.0, release: 1.\ May
  1993 was used. All comments refer to that version.} offers the first
two devices, macros and something resembling lexical rules, the
grammar was written in ALE, and then automatically
translated\footnote{A technical note on the translation: Since in
  Troll everything is encoded in a feature structure data structure, a
  few special types (e.g.~\typ{ps-rule}) and a type for every
  definite clause together with its appropriate argument types are
  added to the translated grammar. Encoding definite clauses as
  feature structures enables the grammar writer to restrict the
  arguments of each definite clause type by specifying the signature.
  This enables the system to do more static type checking.} to Troll.
At the time this work was carried out, neither of the two systems
offered a real debugger; so debugging of the grammar was done the old
way: by thoroughly studying the grammar and massive testing.

\section{The HPSG ontology: A sketch of the logical foundations}
\label{ontology-section}

Before going through the ingredients of an HPSG grammar and the
possibilities for formalizing them in the various architectures, let
us sketch the ontological and logical basis of the architectures we
will mostly be dealing with: \hpsgII, ALE, and Troll.

This sketch seems worth making here, since the setup has direct
consequences on the possibilities in and the extendibility of the
logic based systems.  A lot of the differences between the systems
discussed later on can be traced back to the way the syntax and
semantics of the logic behind the system is set up.  Additionally this
short introduction to the logical basis allows us to introduce the
terminology needed later on (and its diverging use in the various
architectures).

The HPSG formalism as defined in \hpsgII\ is at the basis of the
linguistic analysis developed within the HPSG paradigm over the last
several years.\footnote{The setup of \citeN{hpsg1} differs in
  several respects from that of \hpsgII.} We therefore start by
looking at the ontological setup described in the introduction of
\hpsgII:

\begin{quote}
  In any mathematical theory about an empirical domain, the phenomena
  of interest are {\it modelled\/} by mathematical structures, certain
  aspects of which are conventionally understood as corresponding to
  observables of the domain.  The theory itself does not talk directly
  about the empirical phenomena; instead, it talks about, or is {\it
    interpreted by\/}, the modelling structures.  Thus the predictive
  power of the theory arises from the conventional correspondence
  between the model and the empirical domain. (\hpsgII, page 6)
\end{quote}

In other words, the first step of every scientific process is the
modelling of the domain.  Models\footnote{Note that the term {\it
    model\/} has two usages: A mathematical entity which stands in a
  one-to-one relation to an object of the domain is called a model of
  that object. The second usage, which is avoided in this paper to
  prevent confusion, is to speak of the model of a theory as the set
  of ``things'' which are admitted by the theory.  This double use of model
  is especially confusing, since the ``things'' in the formulation of
  the second usage, are often the models of objects in the first sense of
  the word.} of empirically observable objects are established to
capture the relevant properties.  Theories then make reference to
these models.  This opens up three questions: What do the mathematical
structures used as models for HPSG theories look like?  How are they
related to the linguistic objects?  and How is the linguistic theory
put to work on the models?

\begin{quote}
  In HPSG, the modelling domain \dots\ is a system of {\it sorted
    feature structures\/}, which are intended to stand in a one-to-one
  relation with types of natural language expressions and their
  subparts.  The role of the linguistic theory is to give a precise
  specification of which feature structures are to be considered
  admissible; the types of linguistic entities which correspond to the
  admissible feature structures constitute the predictions of the
  theory. (\hpsgII, page 8)
\end{quote}

The sorted feature structures chosen as models are graph theoretic
entities. Pollard and Sag require them to be {\it totally
  well-typed\/}\footnote{{\it Totally well-typed\/} means that a)
  ``what attribute labels can appear in a feature structure is
  determined by its sort; this fact is the reflection within the model
  of the fact that what attributes \dots\ an empirical object has
  depends on its ontological category.'' and b) ``every feature that
  is appropriate for the sort assigned to that node is actually
  present.''(\hpsgII, p.\ 18)} and {\it sort-resolved\/}\footnote{``A
  feature structure is {\it sort-resolved\/} provided every node is
  assigned a sort label that is \dots\ most specific in the sort
  ordering.''(\hpsgII, p.\ 18)}.  The reason for this is that feature
structures that serve as models of linguistic entities are required to
be total models of the objects that they represent.\footnote{It is not
  uncontroversial whether total models of complete linguistic objects
  are what is needed for linguistics. Problems for such total models
  arise for example in the analysis of coordination (cf.\ the weak
  coordination principle and fn.\ 14 on p.\ 400 of \hpsgII).} To get
the full picture of the setup proposed, we need to answer two more
questions:

How is a theory formulated? The theory is formulated using a specific
description language.\footnote{In \hpsgII\ the term {\it
    attribute-value matrix (AVM)\/} is used instead of {\it
    description\/}. For easier comparison of the different setups, we
  will only use the term {\it description\/}. Furthermore, {\it
    attribute\/} and {\it feature\/} are used synonymously throughout
  the paper. The same is true for {\it type\/} and {\it sort\/},
  except in the use of {\it typed feature structure\/} (cf.\ 
  \citeN{Carpenter92}) vs.\ {\it sorted feature structure\/} (cf.\ 
  \hpsgII).} The description language allows us to express that a
feature structure has to have a certain type, that the value of an
attribute of a feature structure has to have a certain type, or that
the values of two attributes of a feature structure are {\it token
  identical\/}\footnote{The values of two attributes are {\it token
    identical\/} iff the two paths point to the same node. The other
  common identity notion is {\it type identity\/}: Two feature
  structures are {\it type identical\/} iff they are of the same type,
  the same attributes are defined on both feature structures, and the
  feature structures which are the values of the attributes are {\it
    type identical\/}.}. Additionally conjunction, disjunction and
full negation are defined as operators on the descriptions.

What does it mean for a theory to specify admissible feature
structures?  A theory consists of a set of descriptions which are
interpreted as being true or false of a feature structure in the
domain. A feature structure is admissible with respect to a certain
theory iff it satisfies each of the descriptions in the theory and so
does each of its substructures.  The descriptions which make up the
theory are also called {\it constraints\/}, since these descriptions
constrain the set of feature structures which are admissible with
respect to the theory compared to the domain of feature structures
specified by the signature. Note that the term constraint has been
used for many different purposes. In this paper ``constraint'' will
only be used for ``description which is part of the theory''.

The setup of \hpsgII\ is summed up in figure~\ref{hpsgII-setup}.  The
usual symbols for description language operators are used: conjunction
(\und), disjunction (\oder), negation (\nicht), and implication
(\impl).  Additionally, type assignment, path equality, and path
inequality are noted as ``\(\sim\)'', ``\(=\)'', and ``\(\neq\)'',
respectively.

\myfig{hpsgII-setup}{The setup desired in
  \hpsgII}{\hspace*{-1.46em}\mbox{\parbox[t]{8.1em}{\centering
      Descriptions \\[-0.8ex] 
      \(\scriptsize(\tassign,\gleich\;\mid\;\und,\oder,\nicht)\)}\interpret
    \hs Sorted feature structures \oneOne Abstract linguistic object}}

Several logics have been proposed to provide the formal foundations
for this setup. Basically there are two families of logics dealing
with this task: the {\it Kasper-Rounds\/} logics\footnote{Cf.\ 
  \citeN{Rounds&Kasper86}, \citeN{Moshier&Rounds87}, and
  \citeN{Carpenter92}.} and the {\it Attribute-Value\/}
logics\footnote{Cf.~\citeN{Johnson88}, \citeN{Smolka88}, and
  \citeN{King89}.}. Two representatives of these families are of
particular interest for the following discussion, since they also have
been used to construct computational systems for the implementation of
HPSG grammars: the Kasper-Rounds logic defined in
\citeN{Carpenter92} on which the ALE system is based, and the
Attribute-Value logic of \citeN{King89} which underlies the Troll
system. Figure~\ref{carp-setup} shows the setup proposed in
\citeN{Carpenter92}. The descriptions of \citeN{Carpenter92}
describe {\it typed feature structures\/}, which model {\it partial
  information\/}.

\myfig{carp-setup}{The setup of \citeN{Carpenter92}}{
  \hspace*{-1.2em}\mbox{\parbox[t]{8.1em}{\centering Descriptions \\ [-0.8ex] 
      \(\scriptsize(\tassign,\gleich,\ungleich\;\mid\;\und,\oder)\)}
    \interpret \hs Typed feature structures \oneOne \hs Partial
    information}}

This setup differs from the \hpsgII\ desideratum described above.
\citeN{Carpenter92} uses typed feature structures to model {\it
  partial information\/}. Presumably there is a further level, in
which the partial information is related to the linguistic objects,
which is left undefined. Since the typed feature structures model
partial information they cannot be total representations like the
sorted feature structures of \hpsgII.  The typed feature structures
used as models in \citeN{Carpenter92} are therefore not required to
be totally well-typed or sort-resolved.  In the \hpsgII\ setup
``partial information'' plays no role; the linguistic objects
themselves are modelled by the sorted feature structures. This
difference has consequences for the description language of
\citeN{Carpenter92}.  \citeN{Moshier&Rounds87} show that in the
setup of a Kasper-Rounds logic, full classical negation as part of the
description language destroys subsumption monotonicity on typed
feature structures which Carpenter and others claim has to be upheld
if feature structures are to model partial information. In the
discussion of recursive type constraints in chapter 15 of
\citeN{Carpenter92} (p.\ 233), Carpenter states that even for the
formulation of implicative constraints with type antecedents (i.e.\ 
type negation) subsumption monotonicity cannot be
achieved.\footnote{Note that the theories in an \hpsgII\ architecture
  are formulated using implications on types; in fact, even stronger
  implicative statements with complex antecedents are used. We will
  see on pp.~\pageref{type2encoding} of section~\ref{group1-section}
  how certain stronger implicative statements can be reformulated as
  implications with type antecedents by encoding additional
  information in the signature.} The description language of
\citeN{Carpenter92} therefore only contains path-inequations, a
weaker form of negation.

\myfig{king-setup}{The setup of \citeN{King89}}{
  \hspace*{-1.1em}\mbox{\parbox[t]{8.1em}{\centering Descriptions \\ [-0.8ex] 
      \(\scriptsize(\tassign,\gleich\;\mid\;\und,\oder,\nicht)\)} \interpret
    \hs Concrete linguistic objects}}

The setup of \citeN{King89}, displayed in figure~\ref{king-setup},
at first sight also differs from the \hpsgII\ desideratum. The most
apparent difference is the missing model level. The descriptions are
directly interpreted as being true or false of linguistic objects, not
of feature structures which stand in a one-to-one relation with
linguistic objects, as in \hpsgII. Apart from the philosophical
consequences (which will not be discussed here) this gap between
\hpsgII\ and \citeN{King89} can be bridged. Since the models of
linguistic objects in the first approach and the linguistic objects in
the second are both total (i.e.\ the sorted feature structures of
Pollard and Sag are defined that way and the linguistic objects of
King are total since they are objects) there are no formal
consequences to the dropping of the modelling level in
\citeN{King89}.  As proof of this \citeN{King94b} shows that a
modelling level can be introduced without changing the rest of the
logic. The second difference between the \hpsgII\ desideratum and the
King logic is that \hpsgII\ talks about {\it abstract\/} linguistic
objects while the King setup uses {\it concrete\/} linguistic
objects\footnote{Abstract linguistic objects are also called
  linguistic ``types'', and the term linguistic ``tokens'' is used for
  concrete linguistic objects. This usage of the term ``type'' has
  nothing to do with the types as part of the signature of a typed
  feature logic. In order to avoid confusion, the terms ``types'' and
  ``tokens'' meaning abstract and concrete will not be used in this
  paper.}. \citeN{King94b} shows how this gap can be bridged as well.
The logic proposed in \citeN{King89} therefore can be used to
provide the setup desired in \hpsgII.

The difference between the two logics becomes clearer, when we take a
closer look at how descriptions are interpreted in each setup.

\myfig{king-interpretation}{Set theoretic interpretation of
  description language expressions in \citeN{King89}}{
\begin{tabular}[t]{ccc}
  {\bf Syntax} & \interpret & {\bf Semantics}\\[1ex]
  conjunction of descriptions & -- & set intersection \\ 
  disjunction of descriptions & -- & set union \\
  negation of descriptions & -- & set complement
\end{tabular}}

In \citeN{King89} descriptions are given a set theoretic
interpretation: The interpretation of a description is a set of
objects. Conjunction, disjunction, and negation as operations on
descriptions are interpreted as set intersection, set union, and set
complement, respectively. In such a set theoretic setup, an object
satisfies a description iff it is in the denotation of that
description.

\myfig{carp-interpretation}{Interpretation of description language
  expressions in \citeN{Carpenter92}}{
\begin{tabular}[t]{ccc}
  {\bf Syntax} & \interpret & {\bf Semantics}\\[1ex]
  conjunction of descriptions & -- & unification\\
  disjunction of descriptions & -- & eliminated by
    lifting \\[-0.7ex] 
& & disjunction to top level
\end{tabular}}

In \citeN{Carpenter92}, the interpretation of a description is a set
of typed feature structures. Conjunction of descriptions is
interpreted as unification of feature structures and disjunction is
eliminated by lifting it to the top level. The satisfaction relation
is defined directly between feature structures and descriptions
without using set theoretic denotations (cf.\ p.\ 53 of
\citeN{Carpenter92}).

Now that we have introduced the two logics, we can turn to the
computational systems which are based on them. The ALE system uses the
setup and the description language of \citeN{Carpenter92}.
Additionally feature structures in ALE are required to be
totally-well-typed. The main difference between \citeN{Carpenter92}
and the ALE system is that theories in ALE are expressed using a
relational extension\footnote{For a discussion and the formal
  definition of a relational extension of a constraint language cf.\ 
  \citeN{Jaffar&Lassez87} and \citeN{Hoehfeld&Smolka88}.} of the
description language, which is outside of the defined feature
logic.\footnote{The different possible ways of expressing a theory are
  discussed in section~\ref{constraint-section}.}

The \citeN{King89} logic supplies the Troll system with an
interpretation of a signature and a description language which is
adequate for \hpsgII.  However, there are some differences between the
logic and the computational system: Just as in the ALE system, a
relational extension of the description language is used to express
linguistic theories. Also, currently only the conjunctive part of the
description language is part of Troll. As supplement to the King
setup, typed feature structures are introduced as normal form
descriptions. Note that the typed feature structures here are part of
the syntax, not of the semantics as in the Carpenter approach.

There are several consequences of the theoretic and computational
setups just introduced for some commonly used terminology.  When two
descriptions are conjoined in Troll, {\it unification\/} is used to
calculate a normal form description, the denotation of which is the
intersection of the denotations of the descriptions conjoined; i.e.\ 
unification is an operation on the syntax and plays no theoretic role.
In ALE, unification is the interpretation of the conjunction of
descriptions, i.e.\ it is the semantics of a syntactic operation.

The same division into syntax and semantics applies to {\it
  subsumption\/}. In ALE, subsumption is a relation that makes
statements about the complexity of the feature structures serving as
semantic representation. If we try to add subsumption to the King
setup (or that of \hpsgII), we note that subsumption does not make
sense as relation on objects (or the sorted feature structures) which
are on the semantic side, since these are total representations. The
only place for subsumption in the King setup therefore is on the
syntactic side as a relation on descriptions. The interpretation of
the subsumption relation on descriptions is the subset relation on the
denotations of the descriptions. Note, however, that the subsumption
relation is no proper part of the description language of the logic of
\citeN{King89}.  Summing up the last two paragraphs, neither
unification nor checking for subsumption is part of the description
languages of the two logical setups. In the discussion of the lexical
rule mechanism in section~\ref{lex-rules} we will come back to this.

In the rest of the paper we will assume the ontological setup and
terminology of \citeN{King89}. The following feature logic
terminology is used throughout the paper: \(\top\) is the least
constrained type denoting all objects; \(\bot\) is the inconsistent
type denoting the empty set.  {\it Minimal types\/} are the types
which have no subtypes except for \(\bot\), e.g.\ the most specific
types, sometimes also called {\it varieties\/} or {\it species\/}.
Since nothing of theoretical interest relies on the feature structures
as normal form descriptions, feature structures will not appear on
stage in the rest of the paper.  Only when specifically speaking about
the logic of \citeN{Carpenter92} or the ALE system, will the
Carpenter setup and its terminology be used.

\section{The HPSG architecture: A linguist's view on the feature logic toolkit}
\label{architecture-section}

After the ontological primer, we can now go on to discuss the HPSG
architecture based on this setup.  An HPSG theory consists of two
ingredients: the declaration of the domain of linguistic objects in a
signature (consisting of the type hierarchy and the appropriateness
conditions) and the formulation of constraints on that
domain\footnote{The different linguistic motivations behind the
  constraints (e.g.\ grammatical principles, lexical information) and
  the possible ways for expressing these are discussed in section~\ref{ling-constraint-section}.}.  The perspective taken in the
following discussion is always that of a linguist wanting to know what
consequences each formal setup has on the formulation and
implementation of an HPSG-style grammar. The reader interested in the
mathematical definition and properties of the architectures is
referred to the feature logic literature, in particular
\citeN{King89}, \citeN{Carpenter92}, \citeN{King94b}, and
\citeN{Goetz94}.

\subsection{Declaring the domain of a grammar}
\label{signature-section}

From the viewpoint of linguistics, the signature introduces the domain
of objects the linguist wants to talk about. The theory the linguist
proposes for a natural language distinguishes between those objects in
the denotation which are part of the natural language, and those which
are not. First, we will take a look at the signature. The
possibilities for writing theories are dealt with in the next section.

The signature serves as the data structure declaration for the
linguistic theory. It consists of a type hierarchy and appropriateness
conditions. The type hierarchy introduces the classes of objects the
grammar makes use of, allowing us to refer to the classes by using the
types as names. The appropriateness declarations define which class of
objects has which attributes with which values. 

At which types attributes can be introduced and the exact
interpretation of the type hierarchy is subject of debate. Two
interpretations of a type hierarchy are common: ALE is based on an
open world interpretation, while HPSG theory and Troll use a closed
world interpretation.

Briefly said, a closed world interpretation makes two assumptions:
every object in the denotation of a non-minimal type is also described
by at least one of its subtypes; and every object is of exactly one
minimal type. An open world interpretation makes no such assumptions.

Once we commit ourselves to a closed world interpretation, more
syntactic inferences can be made. While both interpretations allow the
inference that appropriateness information present on a type gets
inherited to its subtypes, we can now additionally infer the
appropriateness specifications on a type from the information present
on its subtypes.

In the implementation, the information resulting from these additional
inferences allows us to specify less information by hand and enables
us to express constraints on the interdependence of attribute
specifications.\footnote{Note the correspondence with the so-called
  {\it feature cooccurrence restrictions\/} of \citeN{gkps}. Cf.\ 
  \citeN{Gerdemann&King93} and \citeN{Gerdemann&King94} for a
  detailed investigation of this issue.} Besides the gained expressive
power, this leads to syntactic detection of more errors as well as an
increase in efficiency. 

The following example from the implemented grammar illustrates the
additional expressive power gained in a closed world interpretation.
We want to capture that non-finite verbs are never inverted.

\myfig{intended-inf}{The implication to be captured}{
\avm{VFORM & \nfin} \impl \avm{INV &\minus}}

It is not possible to directly encode the intended relationship
between VFORM and INV in the signature of the feature logics
introduced.  However, under a closed world interpretation the
regularity can be encoded in a signature with a slightly modified type
hierarchy.\footnote{The method used here for encoding implicative
  statements with complex descriptions as antecedent in the signature
  is discussed in detail in section~\ref{group1-section}.} Figure
\ref{verb-hierarchy} shows the type hierarchy below \typ{verb}. Two
new subtypes are introduced.\footnote{Once the subtypes of \typ{verb}
  are introduced, the original VFORM attributes could be omitted
  without loss of information. However, this would change the original
  type hierarchy (with its linguistic motivation) even more, making it
  necessary to reformulate every reference to the verb-form in the
  whole grammar.  In any case, this issue does not change anything
  regarding the fact that the implication (d) in figure~\ref{cw-inferences} will only be obtained under a closed world
  interpretation.}

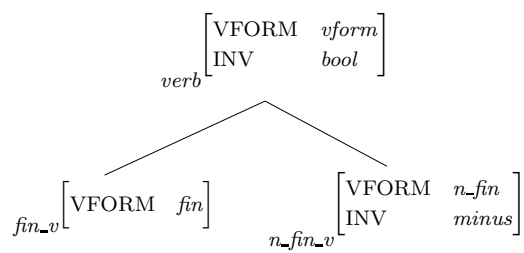
\myfig{verb-hierarchy}{The subtypes of \typ{verb} (a subtype of
  \typ{head})}{
\begin{tabular}[t]{cc}
\mcc{2}{\n{m}{\avmt{verb}{VFORM & \typ{vform}\\INV & \bool}}} \\[8ex]
\n{d1}{\avmt{\finv}{VFORM & \fin}} & \n{d2}{\avmt{\nfinv}{VFORM &
\nfin\\INV & \minus}} \nc{m}{d1}\nc{m}{d2}
\end{tabular}\vspace{1ex}
}

\renewcommand{\arraystretch}{1} 

Under a closed world interpretation the following inferences hold.

\myfig{cw-inferences}{Some inferences under a closed world
interpretation}{
\begin{tabular}[t]{rlcl}
(a) & \finv & \(\leftrightarrow\) & \avm{VFORM & \fin} \\
(b) & \nfinv & \(\leftrightarrow\) & \avm{VFORM & \nfin} \\
(c) & \nfinv & \(\to\) & \avm{INV & \minus} \\
(d) & \avm{INV & \plus} & \(\to\) & \avm{VFORM & \fin} \\ 
(e) & \avm{VFORM & \nfin} & \(\to\) & \avm{INV & \minus}
\end{tabular}}

The originally desired implication repeated as (e) is deduced from (b)
and (c).

Under an open world interpretation only the inferences in figure
\ref{ow-inferences} are made:

\myfig{ow-inferences}{The inferences under open world
interpretation}{
\begin{tabular}[t]{rlcl}
(a) & \finv & \(\to\) & \avm{VFORM & \fin} \\
(b) & \nfinv & \(\to\) & \avm{VFORM & \nfin} \\
(c) & \nfinv & \(\to\) & \avm{INV & \minus}
\end{tabular}}

Compared to the inferences made under a closed world interpretation,
under an open world interpretation the ``\(\gets\)'' direction in (a)
and (b) is missing .  Therefore the originally desired implication
(figure~\ref{intended-inf}) cannot be deduced. Additionally the
inference (d) of figure~\ref{cw-inferences} is missing. To still get
the desired results one therefore has to ensure the ``\(\gets\)''
direction of (a) and (b) by hand; i.e.\ additional specifications on
the lexical entries and the principles are necessary.  The same is
true for the implication (d).  Under an open world interpretation the
full set of inferences of figure~\ref{cw-inferences} (which include
the originally desired implication of figure~\ref{intended-inf})
therefore only hold for the descriptions bearing the additional
specifications; i.e.\ under an open world interpretation it is not
possible to ensure that the full set of inferences will always be made
for every object.  Regarding the theoretical level, this can lead to
incorrect analysis.  In the computational systems based on an open
world interpretation (e.g.\ ALE) this has the effect that the desired
inferences will not be made for feature structures resulting in
processing. The missing inferences in such computational systems
therefore result in possibly incorrect, but in any case slower
analysis.

The second variation in the concept of a signature concerns the
appropriateness conditions. The feature introduction condition (FIC)
of ALE demands that for every feature there is a unique least specific
type at which this feature is introduced.

In the grammar implemented this turns out to be troublesome. As an
illustration let us look at the signature definition of \typ{head}
objects. Figure~\ref{head-hierarchy} shows the type hierarchy as
motivated by the linguistic theory.\footnote{As before and throughout
  the paper, the most general type is on top. Note that Carpenter's
  hierarchies are drawn the other way around: his most general type is
  at the bottom.}

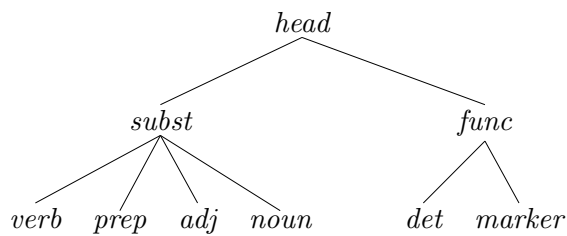
\myfig{head-hierarchy}{The type hierarchy below \typ{head}}{ {\it
    \begin{tabular}[t]{ccccccc} \mcc{7}{\n{1}{head}} \\[5ex]
\mcc{4}{\n{2}{subst}}  & & \mcc{2}{\n{3}{func}} \\[5ex]

\n{4}{verb} & \n{5}{prep} & \n{6}{adj} & \n{7}{noun}& & \hs\n{8}{det} &
\n{9}{marker}

\nc{1}{2}\nc{1}{3} \nc{2}{4}\nc{2}{5}\nc{2}{6}\nc{2}{7}
\nc{3}{8}\nc{3}{9}
\end{tabular}}
}

The problem arises with the introduction of the case and declension
attributes, since CASE and DECL are appropriate for adjectives, nouns,
and determiners, but not for markers and verbs. To satisfy the feature
introduction condition, a new type \typ{adj\_noun\_det} as subtype of
\typ{head} and with subtypes \typ{adj}, \typ{noun}, and \typ{det}
needs to be introduced for which the two attributes CASE and DECL are
appropriate.

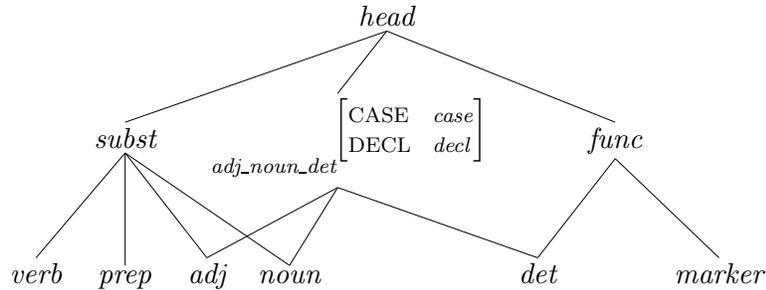
\myfig{head-hierarchyB}{A first step towards satisfying the FIC}{
{\it \begin{tabular}[t]{cccccccc}
\mcc{8}{\n{1}{head}} \\ [5ex]
& \n{2}{subst} & \mcc{3}{\n{3}{\avmt{adj\_noun\_det}{CASE &
        \typ{case} \\ DECL & \typ{decl}}}} & & \n{4}{func} &\\ [8ex]
\n{5}{verb} & \n{6}{prep} & \n{10}{adj} & \n{11}{noun} & &
\n{8}{det} & & \n{9}{marker} \\ [5ex]

\nc{1}{2}\nc{1}{3}\nc{1}{4}
\nc{2}{5}\nc{2}{6}\nc{2}{10}\nc{2}{11}
\nc{3}{10}\nc{3}{11}\nc{3}{8}
\nc{4}{8}\nc{4}{9}
\end{tabular}}
}

However, in the resulting structure in figure~\ref{head-hierarchyB}
the types \typ{subst} and \typ{adj\_noun\_det} do not have a unique
greatest lower bound. This violates the formalization of a type
hierarchy as {\it finite bounded complete partial order\/} (cf.\ 
{Carpenter92}).  Therefore, an additional type \typ{adj\_or\_noun}
needs to be introduced.  Figure~\ref{fic-figure} shows the final type
hierarchy together with the appropriateness information for
\typ{adj\_noun\_det}.\footnote{The type hierarchy actually used in the
  implementation is more complicated still, since a type
  \typ{non\_noun} as subtype of \typ{head} has to be introduced.  This
  is necessary to introduce a type for lists of nominal signs, which
  is needed as constraint on the auxiliary entries and the output of
  the PVP extraction lexical rule in HN.  Cf.\ section~\ref{lex-rules}, p.~\pageref{list-np-type-mention} for a
  discussion.}

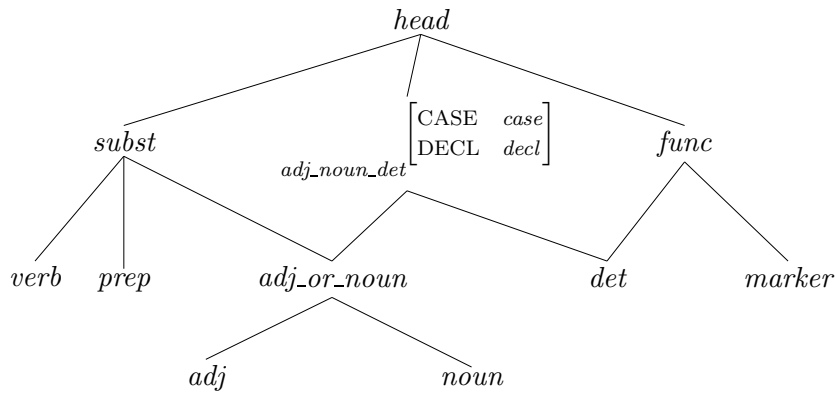
\myfig{fic-figure}{A type hierarchy satisfying the FIC for CASE and
  DECL}{ {\it \begin{tabular}[t]{cccccccc} \mcc{8}{\n{1}{head}} \\ 
    [5ex]

  & \n{2}{subst} & &\mcc{2}{\n{3}{\avmt{adj\_noun\_det}{CASE & \typ{case}
      \\ DECL & \typ{decl}}}} & & \n{4}{func} \\ [8ex]

  \n{5}{verb} & \n{6}{prep} & & \n{7}{adj\_or\_noun} & & \n{8}{det} &
  & \n{9}{marker} \\ [5ex]

& & \n{10}{adj} & & \n{11}{noun}

\nc{1}{2}\nc{1}{3}\nc{1}{4}
\nc{2}{5}\nc{2}{6}\nc{2}{7}
\nc{3}{7}\nc{3}{8}
\nc{4}{8}\nc{4}{9}
\nc{7}{10}\nc{7}{11}
\end{tabular}}
}

The feature introduction condition is part of the
\citeN{Carpenter92} logic and the ALE system. It plays no role in
the logic of \citeN{King89} or the Troll
system.\footnote{\citeN{King&Goetz93} show that the feature
  introduction condition can also be eliminated within the logic of
  \citeN{Carpenter92}.}

\subsection{Constraining the domain}
\label{constraint-section}

Now that we have seen how the domain of linguistic objects is defined
by the signature, we are ready to discuss how the theory, i.e.\ the
constraints making up the grammar can be expressed.  The theory
divides the domain of objects specified by the signature into
admissible ones, which are part of the natural language described, and
those which are not admissible, i.e.\ not part of the natural
language. In linguistic terms the constraints making up the grammar
are the principles (e.g.\ the Immediate Dominance Principle (IDP), the
Head Feature Principle (HFP), the Nonlocal Feature Principle (NFP),
the Semantics Principle (SemP)) and the specification of the lexicon.
The architectures of \hpsgII\ and the various systems differ
considerably regarding the question of how the constraints are
expressed.  Basically there are two main groups: those description
languages directly constraining the domain of linguistic objects, and
those in which relations on linguistic objects are expressed.

\subsubsection{Group~1: Directly constraining the domain of linguistic objects}
\label{group1-section}

In \hpsgII\ and the TFS system the grammatical constraints are expressed
as statements about the domain of linguistic objects. The simplest
implicative form (henceforth called {\it type definition\/}) is
depicted in figure~\ref{type-definition}. It has the following
interpretation: if something is of type \typ{t}, then it has to
satisfy the description $D$.

\myfig{type-definition}{The type definition: a simple implicative
  constraint}{\(t \hspace{1em} \impl \hspace{1em} D\)}

In addition to the type definitions, \hpsgII\ also uses a stronger
implicative statement of the form displayed in figure
\ref{object-definition}. 

\myfig{object-definition}{The object definition: an implicative
  statement with possibly complex antecedent}{\(C \hspace{1em} \impl
  \hspace{1em} D\)}

The intended interpretation is that if something satisfies the
possibly complex description \(C\) it also has to satisfy \(D\). We
will call the implicative constraints (including the type definitions)
{\em object definitions\/}.  In writing a grammar one usually wants to
impose constraints on a certain group of objects (those in the
denotation of the antecedent) and not on all objects (in which case a
non-implicative statement would do). Therefore, object definitions are
the main building blocks of linguistic theories in the HPSG
architecture.  As an example, take the HFP as formulated in the
appendix of \hpsgII:

\myfig{hfp-avm}{The Head-Feature Principle of
  \hpsgII}{\avmt{phrase}{DTRS & \hspace{-2ex} \typ{headed-struc}}
  \hspace{1em} \impl \hspace{0.9em} \avm{SYNSEM\|LOC\|CAT\|HEAD &
    \idx{1} \\DTRS\|HEAD-DTR\|SYNSEM\|LOC\|CAT\|HEAD & \idx{1}}}

The complex description \avmt{phrase}{DTRS & headed-struc} is the
antecedent of the implication. Note that using the type \typ{phrase}
as antecedent is not enough, since the HFP is only supposed to apply
to phrases dominating headed structures. The HFP constraint demands
that all objects of type \typ{phrase} which have a DTRS attribute with
a \typ{headed-struc} object as value, also have to be described by the
consequent, i.e.\ satisfy the description that the HEAD values of the
object and that of its head daughter are token-identical.

\mysection{A closer look at the antecedents of implicative
  constraints}

To get an idea of the complexity of the description language, we need to
distinguish between different kinds of antecedents of implicative
constraints.  Figure~\ref{complexity-hierarchy} displays three classes
of descriptions, which can function as antecedents of implicative
statements in the order of increasing complexity.

\myfig{complexity-hierarchy}{Three classes of descriptions}{
\begin{itemize} 
\item[1.] simple type description (e.g.~\typ{t})
\item[2.] complex description with type assignments (e.g.~\avmt{t}{X &
    \avmt{u}{Y & v \\ Z & w}})
\item[3.] complex description with type assignments and path
  equalities (e.g.~\avmt{t}{X & \idx{1}\typ{u} \\ Y & \idx{1}})
\end{itemize}}

The simplest implicative statements are the type definitions, which
only allow types as antecedent (class 1). One step up in complexity,
complex feature structures which specify no structure sharing can be
used as antecedents (class 2). Finally, in the most complex type of
implications the antecedents are allowed to specify structure sharing
information as well (class 3).

The implicative statements in \hpsgII\ are mostly of class 2. However
none of the computational systems allow the user to specify complex
antecedents; only type definitions are supported.\footnote{This also
  applies to the systems in which a theory is expressed on the
  relational level (the group~2 setups) since the method described in
  section~\ref{group2-section} for translating theories which are
  expressed as constraints on the domain (group~1) into theories which
  are expressed on the relational level (group~2) is restricted to
  type definitions as well.  Therefore, until other general techniques
  for expressing HPSGII-style theories on the relational level are
  defined which might allow us to express implicative constraints with
  complex antecedents, both computational setups are restricted to
  type definitions.} Because of this discrepancy in expressive power,
it is interesting to note that the implicative statements with complex
antecedents used in \hpsgII\ can be expressed as type definitions by
modifying the signature, i.e.\ a new type can be introduced so that it
has the same denotation as a given class 2 description. To see how
this encoding works, let us first take a closer look at the
appropriateness conditions (ACs).

In section~\ref{signature-section} we had characterized ACs as part of
the data structure declaration, the declaration of the domain in more
logic based terminology.  This means, the ACs take part in defining
the domain over which the theory is formulated. The objects not
satisfying the ACs are not in the domain. On the other hand, ACs can
also be understood as rather weak constraints on the domain of objects
as defined by the type hierarchy alone.\footnote{Note that ACs are
  inherited from a type to its subtypes, allowing generalizations to
  be expressed in a nice hierarchical way independent of how the
  theory is formulated.} They restrict the values of the immediate
attributes of a type without specifying structure sharing. Under this
interpretation, the domain over which the theory makes statements also
contains objects which do {\it not\/} satisfy the AC. However, while
the one-level-only condition does not restrict the set of descriptions
which can be encoded in the signature definition of a type, the
no-structure-sharing condition is a real restriction. Descriptions
specifying structure sharing cannot be encoded in a type by modifying
the signature only. To encode a class 3 description in a type, one
therefore additionally has to specify a constraint on the new type as
part of the theory in order to express the structure sharing
conditions on its attributes.\footnote{So far no HPSG theory to our
  knowledge makes use of class 3 antecedents for implicative
  statements. This means that currently only the class 2 antecedents,
  which can be encoded without manipulating the theory, i.e.\ by
  changing the signature only, are linguistically motivated.}

\label{type2encoding} 

Encoding the information of a class 2 description \(D\) in the
signature works in the following way: A new type is introduced as
subtype of the type of D.\footnote{A complication arises, if \(D\)
  describes objects of different types. In this case, a new subtype
  has to be introduced for each of the types of object described by
  D.} The type specifications on the first level of attributes of
\(D\) are encoded in the ACs of the new type.  If \(D\) contains
descriptions of further levels of attributes, new subtypes have to be
introduced to mediate the information.  Note that this introduction of
additional types for each additional level quickly leads to an
explosion of the type hierarchy.

To illustrate this mechanism let us look at an example from
HN.\footnote{For this and the following examples I assume the
  signature of HN and the grammar implemented. It closely resembles
  the one in the appendix of \hpsgII\ with the modifications of
  Chapter 9.  However, note that all valence attributes (e.g.\ COMPS)
  take a list of signs as their value, not a list of synsems. The
  order of the valence lists of HN is inverted in the grammar
  implemented, i.e.\ valence lists look like {\it
    \lstI{most-oblique-element \ldots\ least-oblique-element}.\/} This
  order is used in the examples of this paper as well. A discussion of
  the motivation behind this change is given in section~\ref{valence-encoding}.} In HN every auxiliary is supposed to bear
an argument raising specification, which in its basic form is shown in
figure~\ref{arg-raise}.

\myfig{arg-raise}{Argument raising specification on a non-finite
  auxiliary (\typ{valence} value shown)}{\avm{COMPS &
    \lst{\avm{SYNSEM\|LOC\|CAT & \avm{HEAD & \typ{verb} \\ VAL\|COMPS &
          \idx{1}}}}{\idx{1}}}}

There are two options for encoding this in the grammar. Since we are
dealing with lexical specifications, the methods for expressing
generalizations over classes of lexical entries could be used. These
are dealt with in section~\ref{lexicon}. The other possibility for
attaching the argument raising specification in figure~\ref{arg-raise}
to auxiliaries is to use the general mechanism for expressing
grammatical constraints. To do this, the following object definition
can be added to the grammar:

\myfig{obj-def-arg-raise}{An object definition to attach the
  argument raising
  specification}{\begin{flushleft}\hspace{2em}\avmt{word}{SYNSEM\|LOC\|CAT\|HEAD
    & \avmt{verb}{AUX & \plus}} \hspace{1em} \impl \\[0.4ex]
\end{flushleft}\hspace{3em}\avm{SYNSEM\|LOC\|CAT\|VAL\|COMPS & 
  \lst{\avm{SYNSEM\|LOC\|CAT & \avm{HEAD & \typ{verb} \\ VAL\|COMPS &
        \idx{1}}}}{\idx{1}}}\vspace{1ex}}

\vspace{1ex}

To rewrite this object definition as type definition, a new subtype of
\typ{word} (call it \typ{aux-word}) needs to be introduced. We want it
to have the same denotation as the complex antecedent of the
implication in figure~\ref{obj-def-arg-raise}. Assuming the signature
definitions given in the appendix of \hpsgII\ as basis, under a closed
world interpretation the following flood of signature changes and new
definitions is necessary to achieve that.

\vspace{1ex}

\newlength{\mylength}

\setlength{\mylength}{-1.7ex}

\myfigPB{type-def-aux-word}{Additional signature definitions to
  introduce the type \typ{aux-word}}{
\begin{itemize}
\item[a)] \(\typ{word} \hspace{0.6ex} >
  \hspace{0.3ex} \typ{non-aux-word}, \; \typ{aux-word}.\)
\begin{itemize}
\vspace*{\mylength}
\item[i.] \avmt{non-aux-word}{SYNSEM & \typ{non-aux-synsem}}.
\vspace*{-1.5ex}
\item[ii.] \avmt{aux-word}{SYNSEM & \typ{aux-synsem}}.
\end{itemize}
\item[b)] \(\typ{synsem} \hspace{0.6ex} > \hspace{0.3ex}
  \typ{non-aux-synsem}, \; \typ{aux-synsem}.\)
\begin{itemize}
\vspace*{\mylength}
\item[i.] \avmt{non-aux-synsem}{LOC & \typ{non-aux-loc}}.
\vspace*{-1.5ex}
\item[ii.] \avmt{aux-synsem}{LOC & \typ{aux-loc}}.
\end{itemize}
\item[c)] \(\typ{loc} \hspace{0.6ex} > \hspace{0.3ex}
  \typ{non-aux-loc}, \; \typ{aux-loc}.\)
\begin{itemize}
\vspace*{\mylength}
\item[i.] \avmt{non-aux-loc}{CAT & \typ{non-aux-cat}}.
\vspace*{-1.5ex}
\item[ii.] \avmt{aux-loc}{CAT & \typ{aux-cat}}.
\end{itemize}
\item[d)] \(\typ{cat} \hspace{0.6ex} > \hspace{0.3ex}
  \typ{non-aux-cat}, \; \typ{aux-cat}.\)
\begin{itemize}
\vspace*{\mylength}
\item[i.] \avmt{non-aux-cat}{HEAD & \typ{non-aux-head}}.
\vspace*{-1.5ex}
\item[ii.] \avmt{aux-cat}{HEAD & \typ{aux-verb}}.
\end{itemize}
\item[e)] \(\typ{head} \hspace{0.6ex} > \hspace{0.3ex} \typ{subst}, \;
  \typ{non-aux-head}.\)
\item[f)] \(\typ{subst} \hspace{0.6ex} > \hspace{0.3ex} \typ{verb}, \;
  \typ{non-aux-subst}.\)
\item[g)] \(\typ{non-aux-head} \hspace{0.6ex} > \hspace{0.3ex}
  \typ{non-aux-subst} , \; \typ{func}.\)
\item[h)] \(\typ{non-aux-subst} \hspace{0.6ex} > \hspace{0.3ex}
  \typ{non-aux-verb} , \; \typ{prep}, \; \typ{adj}, \; \typ{noun}.\)
\pagebreak
\item[i)] \(\typ{verb} \hspace{0.6ex} > \hspace{0.6ex}
  \typ{non-aux-verb}, \; \typ{aux-verb}.\)
\begin{itemize}
\vspace*{\mylength}
\item[i.] \avmt{non-aux-verb}{AUX & \typ{minus}}.
\vspace*{-1.5ex}
\item[ii.] \avmt{aux-verb}{AUX & \typ{plus}}.
\end{itemize}
\end{itemize}
}

Two subtypes are introduced for each of the types \typ{word},
\typ{synsem}, \typ{loc}, and \typ{cat} in order to mediate the AUX
specification from the \typ{sign}-level to the \typ{head} level. The
hierarchy below \typ{head} is more complicated. For easier
comprehension it is displayed in graphical notation below. The
corresponding type hierarchy of \hpsgII\, from which this new
hierarchy is derived, was already presented in figure
\ref{head-hierarchy}.

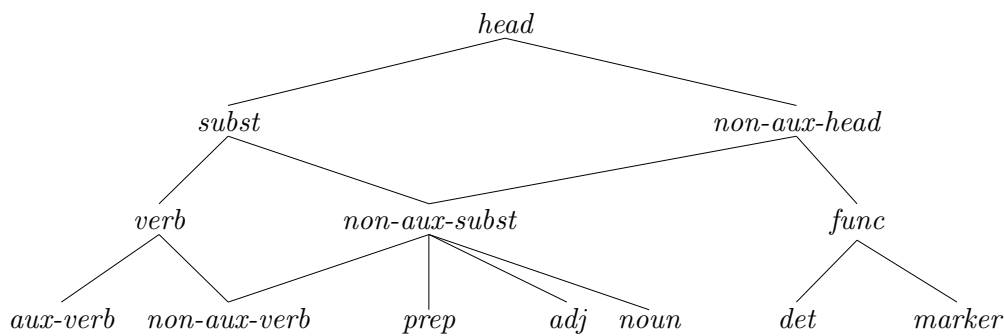
\myfig{aux-encoding-head-figure}{The resulting type hierarchy below \typ{head}}{ {\it
    \begin{tabular}[t]{ccccccc} \mcc{7}{\n{1}{head}} \\ [5ex]
& \n{2}{subst} && & & \n{3}{non-aux-head} & \\ [5ex]
\mcc{2}{\n{4}{verb}} & \n{5}{non-aux-subst} & & & \mcc{2}{\n{6}{func}}\\[5ex]

\n{7}{aux-verb} & \n{8}{non-aux-verb} & \n{9}{prep} & \n{10}{adj} &
\n{11}{noun} & \n{12}{det} & \n{13}{marker} \\[5ex]

\nc{1}{2}\nc{1}{3}
\nc{2}{4}\nc{2}{5}
\nc{3}{5}\nc{3}{6}
\nc{4}{7}\nc{4}{8}
\nc{5}{8}\nc{5}{9}\nc{5}{10}\nc{5}{11}
\nc{6}{12}\nc{6}{13}
\end{tabular}}
}

Clearly these linguistically unmotivated changes to the type hierarchy
in order to lift information to the ``surface'', i.e.\ the top level
of an implication, is nothing that can be asked of a linguist encoding
a grammar.  One could therefore consider eliminating the hierarchical
structure of the models of linguistic objects and return to simpler,
flat models. Until there is linguistic evidence against the
hierarchical structures\footnote{We here ignore the problematic issue
  of lexical hierarchies (cf.\ Chapter 8 of \citeN{hpsg1}). A short
  discussion of these hierarchies is included in section~\ref{lex-classes}.} it is better motivated though, to extend the
computational systems so that complex antecedents can be expressed
directly.  Introducing general negation into the formalism, would
allow any complex antecedent to be expressed directly. However, while
this might turn out to be possible in some underlying logics, losing
the type backbone to the constraints (the type antecedents as
``triggers'' for the implicative constraints) causes significant
computational problems and no system currently provides such an
architecture. In this situation, the technique introduced above for
encoding complex descriptions in types looks like a valuable method
after all. An automated version of the encoding process could be
included in a compilation step applying after the writing of the
grammar. Thus the linguistically unmotivated mutation of the signature
would also be hidden from the eyes of the linguist.

\newpage
\mysection{Recursivity}
\label{recursivity}

After looking at the antecedents of implicative constraints, we now
turn to the consequent of the implications to see where the expressive
power to make recursive statements (necessary to work with lists and
sets) comes from.

TFS allows a more complex version of type definitions, which is
displayed in figure~\ref{tfs-type-def}.

\myfig{tfs-type-def}{The extended type definitions of TFS}{ $t
  \hspace{1em} \impl \hspace{1em} (D \; \& \; E_1 \; \& \; \ldots \;
  \& \; E_n)$}

Just as with simple type definitions, the extended type definition
above is true of those objects of type \typ{t} which are also
described by \(D\). Additionally, each of the new descriptions $E_i$
appended with ``\&''\footnote{The ``\&'' is not to be confused with
  the conjunction ``\und'' of the description language, which is
  interpreted as set intersection of the denotations of the
  conjuncts.} have to be satisfiable.  In other words: there must be
objects in the denotation of each description $E_i$. These extended
type definitions enable us to state recursive definitions, since each
of the objects in the denotation of $E_i$ has to satisfy the
constraints imposed on its type.  If an \(E_i\) describes objects of the
type \typ{t} that is being defined (= the antecedent of the
implication), one obtains a recursive statement. 

\label{relationsIntoDescriptions}
It is possible to make sense of the meta operator operator ``\&''
within the description language. Following \citeN{Ait-Kaci84}, one
can add extra attributes (``junk slots'') to linguistic objects to
``store'' additional objects.  The descriptions which were added using
the extra-logical ``\&'' operator now only have to be added to the
theory as descriptions of the objects residing in the newly introduced
attributes.\footnote{This applies to the TFS system as well. The
  ``\&'' operator in TFS can therefore be seen as surface syntax for
  the user making it unnecessary to worry about introducing additional
  attributes as junk slots to store the additional descriptions which
  one wants to test for satisfiability.} Applying this junk slot
encoding technique to the definition in figure~\ref{tfs-type-def}, we
introduce a list valued attribute STORE as appropriate for type
\typ{t}. The resulting description language encoding of figure
\ref{tfs-type-def} is displayed in figure
\ref{pure-extended-type-def}.

\myfig{pure-extended-type-def}{Extended type definitions expressed using the
  description language only}{\(t \hspace{1em} \impl \hspace{1em}
(D \; \und \; \avm{STORE & \XlstI{\(E_1, \; \ldots, \; E_n\)}})\)}

Beside using the additional descriptions housed in the junk slot
attribute to describe ordinary linguistic objects, one can also use
statements like that in figure~\ref{pure-extended-type-def} to
formulate recursive relations on linguistic objects at the same level
at which the objects are described. The additional junk slot attribute
then also allows calls to relations to be attached to descriptions of
linguistic objects without the use of a meta operator ``\&''. Note
that when a recursively defined relation is used as description of an
object's attribute value, the object will bear the full proof-tree in
the attribute.

As example for the definition of a recursive relation at the level of
the description language, the definition of the standard {\it
  append\/} relation on lists\footnote{The {\it append\/} relation
  specifies the third argument to be the concatenation of the two
  lists passed as first and second argument.} is shown in figure
\ref{type-append}.

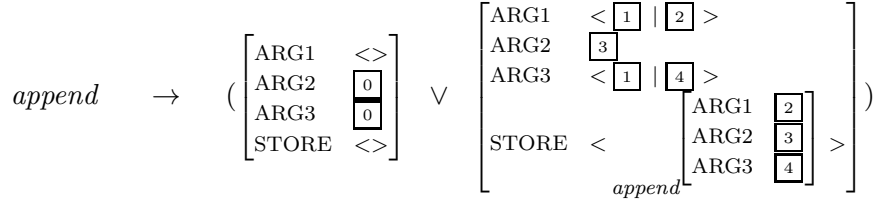
\myfig{type-append}{Defining append objects}{ \typ{append}
  \hspace*{1.2em} \impl \hspace*{1.2em} \((\avm{ARG1 & \Xelst \\ ARG2
    & \idx{0} \\ ARG3 & \idx{0} \\ STORE & \Xelst}\hspace*{0.8em} \oder
  \hspace*{0.8em} \avm{ARG1 & \Xlst{\idx{1}}{\idx{2}} \\ ARG2 &
    \idx{3} \\ ARG3 & \Xlst{\idx{1}}{\idx{4}} \\ STORE &
    \XlstI{\hspace*{-1em}\avmt{append}{ARG1 & \idx{2} \\ ARG2 &
        \idx{3} \\ ARG3 & \idx{4}}}})\)}

The constraint in figure~\ref{type-append} defines what kind of append
objects are admissible.  The description ``\typ{append}'' (i.e.\ a
call to append, possibly as part of some other description) denotes a
set of append objects having their attributes constrained as defined.

While \hpsgII\ is not committed to a certain way of expressing
recursive statements, an alternative possibility often found in the
HPSG literature is to use definite clauses as a relational extension
of the description language. Note that in contrast to the encoding of
relations discussed above, the use of definite clauses described here
is outside of the feature logic and it is not formally defined. The
definitions of the definite clauses are allowed to be recursive,
yielding the expressive power we were looking for. To use the
relations thus defined, the calls to the definite clauses are attached
to the implicative statements just like the additional descriptions
are in TFS.

\myfig{hpsg-constr}{A type definition with attached relational
  constraints\footnotemark}{ $C \hspace{1em} \impl
  \hspace{1em} (D \;\; \& \;\; rel_1(E_{11}, \ldots, E_{1i}) \;\; \&
  \;\; \ldots \;\; \& \;\; rel_n(E_{n1},\ldots,E_{nj}))$.}

\footnotetext{\(C\), \(D\), and the \(E_{xy}\) are descriptions.}

Figure~\ref{hpsg-constr} only shows the use of relations, not their
definition. Contrary to the setup used in TFS, the definition of the
relation itself is specified on a level different from the level of
the grammatical constraints defined in figure~\ref{hpsg-constr}.
Because the definite clause relations are not part of the description
language, the calls to the definite clauses need to be added to the
descriptions using the extra-logical operator ``\&''. Note that for
the same reason it is not possible to include the calls to the
definite clause relations in a description as it was done with the
help of the junk slot technique for the relations formulated within
the description language.  Definite clauses are defined as in PROLOG,
the only difference being that while in PROLOG definite clauses over
first order terms are formulated, here definite clauses over feature
descriptions are used.  Figure~\ref{relation-append} shows the
definition of the {\it append\/} relation as part of the relational
extension of the description language.

\myfig{relation-append}{Defining the append relation}{
\begin{itemize}
\item[a)] append(\Xelst, \idx{1}, \idx{1}).
\item[b)] append(\Xlst{\idx{1}}{\idx{2}}, \idx{3},
  \Xlst{\idx{1}}{\idx{4}}) := append(\idx{2}, \idx{3}, \idx{4}).
\end{itemize}}

Here {\it append\/}, being a three place relation, denotes the set of
triples of list objects, which satisfy the definition given.

This ends the discussion of the first group of setups, in which the
theories directly constrain the domain.

\subsubsection{Group~2: Expressing relations on linguistic objects}
\label{group2-section}

The second group -- represented by systems like ALE, CUF, and Troll --
takes the concept of the relational extension of the description
language just described, and makes it the only way to express the
grammatical constraints of the theory. Relations over objects are
formulated. Unlike in the group~1 setups, the objects cannot be
constrained directly.\footnote{ALE additionally offers the possibility
  to specify type definitions.  Therefore ALE belongs to both group~1
  and group~2 architectures, i.e.\ it can be used to formulate
  grammars in two different ways: by specifying constraints on the
  domain and by writing down relations on objects. However, currently
  the formal status of such hybrid setups is rather unclear. In the
  grammar implemented only the relational level is used to express the
  linguistic theory.}

The result is a setup with a clean distinction between the description
language, and the relational extension of that language: the
description language only serves to describe the objects which can be
arguments of the relations. Figure~\ref{relations} illustrates this
setup.

\myfig{relations}{Two levels of constraints in a definition of the
  relation $rel_0$}{$rel_0(E_{01}, \ldots, E_{0i}) \hspace{0.9em} :-
  \hspace{1em} rel_1(E_{11}, \ldots, E_{1j}), \; \ldots, \;
  rel_n(E_{n1}, \ldots, E_{nk})$}

Note that the question of how the constraints making up the theory are
expressed is a separate issue from the formulation of a signature
discussed in section~\ref{signature-section}.  Both, group~1 style and
group~2 style theories are based on an underlying signature.  However,
as discussed in the previous section, the appropriateness conditions
of the signature can also be seen as expressing weak constraints on a
domain defined by the type hierarchy only. Therefore, even in a group
2 system, i.e.\ in a system in which the theory is expressed
exclusively on the level of a relational extension of the description
language outside of the feature logic defined, some grammatical
constraints can be expressed inside of the logic by encoding them in
the signature.

Definite clauses are not the only relevant way to express relations.
ALE and Troll offer a parser, which allows the formulation of phrase
structure (PS) rules as relations on objects of a local tree. The
phrase structure rules function as backbone to which the grammatical
constraints expressed in definite clauses are attached. Such a setup
was used for the implementation of HN. The method used to encode the
HPSG theory in a phrase structure backbone setup while at the same
time trying to preserve the character of the original theory is
discussed in sections~\ref{struc-licensing} and~\ref{principles}.

The difference between the two groups of architectures is reflected in
the queries which can be formulated in those architectures.  While in
the first group one queries with a description to see whether that
description describes any objects, in the second group one asks
whether a relation holds between objects.

So far we have not yet explained, how an HPSG grammar can be expressed
in a group~2 setup. The \hpsgII\ theory is based on an architecture in
which the domain can be directly constrained by the theory (group~1).
Still, for computational reasons, almost all computational systems use
a group~2 setup.  It is therefore interesting to take a look at how a
setup like the first can be modelled in a system of the second group.

\mysection{Modelling group~1 behavior in a group~2 setup}

What does it need to get the feel of a group~1 setup in a group~2
system? For the linguist wanting to encode a grammar it boils down to
two things: How are the constraints making up the theory specified?
and How is the system queried? Something we need to integrate into a
group~2 setup to get a group-1-like behavior regarding the constraint
specification is the hierarchical organization of constraints.  An
object of a certain type has to satisfy all the constraints on objects
of that type {\it and\/} all the constraints on objects of any of its
supertypes. Regarding the queries, we want to be able to ask if a
description is satisfied by any objects which satisfy the theory.

If we restrict ourselves to implicative constraints with type
antecedents\footnote{This opens up a new area of application for the
  technique for encoding complex antecedents of implicative
  constraints as type antecedents by modifying the signature discussed
  in section~\ref{group1-section} on pp.~\pageref{type2encoding}.}, a
general solution is to introduce three relations for every type. One
relation is needed to encode the constraints defined for the type,
i.e.\ to specify the type and to encode the consequent of the type
definitions for the type which are part of the theory. A second
relation collects all constraints on the type (by calling the first
relation) and its subtypes. Finally a third relation is responsible
for adding the constraints on the supertypes of the type.  Let us
illustrate this method with an example.

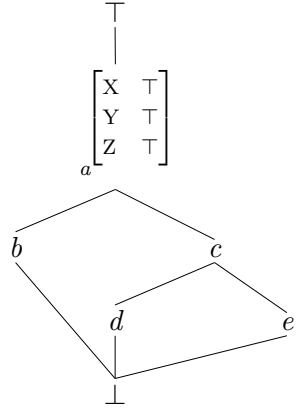
\myfig{modelling-signature}{The example signature}{
\begin{tabular}[t]{cccc}
& \n{top}{\(\top\)}\\[3ex]
& \n{a}{\avmt{a}{X & \(\top\) \\ Y & \(\top\) \\ Z & \(\top\)}}\\[7ex]
\n{b}{\typ{b}} & & \n{c}{\typ{c}} \\[3ex]
& \n{d}{\typ{d}} & & \n{e}{\typ{e}} \\[3ex]
& \n{bot}{\(\bot\)} & & 
\nc{top}{a} \nc{a}{b} \nc{a}{c} \nc{c}{d} \nc{c}{e} \nc{b}{bot}
\nc{d}{bot} \nc{e}{bot}
\end{tabular}}
 
As usual, the signature definition in figure~\ref{modelling-signature}
declares a domain of objects. The theory in figure
\ref{modelling-constraints} specifies which objects are admissible. It
consists of two implicative constraints with types as antecedent.

\myfig{modelling-constraints}{The example theory in a group~1 setup}{
\begin{tabular}[t]{lcl}
\typ{c} & \impl & \avm{X & \idx{1}\avm{Z & b} \\ Y & \idx{1}}\\
\typ{d} & \impl & \avm{Y & \idx{1} \\ Z & \idx{1}}
\end{tabular}}

In figure~\ref{modelling-relations} the group~2 theory with the three
relations for each type is listed.\footnote{A notation reminiscent of
  PROLOG is used for the relational level: disjunction is noted as
  ``;'' and conjunction by ``,''. The full set of three relations per
  type is only displayed to illustrate the general mechanism. In an
  implementation several of the relations defined could be eliminated,
  for example by partially executing the definite clauses at compile
  time.} The relation name subscripts and the names of the lines in
which the relations are defined have the following interpretation:
``h'' stands for the {\it hierarchy\/} of constraints below the type
(including the type itself), ``c'' for the {\it constraint\/} on the
type, and ``t'' for all constraints on the {\it type\/} including
those inherited.

\myfigPB{modelling-relations}{The relations encoding the example
  theory in a group~2 setup}{
\begin{itemize}
\item[1.] \begin{itemize}
              \item[h)] \(\top_{h}\)(\idx{1}) :- a\(_{h}\)(\idx{1}).
           \end{itemize}\vspace{1ex}
\item[2.] \begin{itemize}
              \item[t)] a\(_{t}\)(\idx{1}) :- \(\top_{h}\)( \idx{1}\typ{a} ).
              \item[h)] a\(_{h}\)(\idx{1}) :- a\(_{c}\)(\idx{1}), ( b\(_{h}\)(\idx{1}); c\(_{h}\)(\idx{1}) ).
              \item[c)] a\(_{c}\)(\typ{a}).
          \end{itemize}\vspace{1ex}
\item[3.] \begin{itemize}
              \item[t)] b\(_{t}\)(\idx{1}) :- \(\top_{h}\)( \idx{1}\typ{b} ).
              \item[h)] b\(_{h}\)(\idx{1}) :- b\(_{c}\)(\idx{1}).
              \item[c)] b\(_{c}\)(\typ{b}).
          \end{itemize}\vspace{1ex}
\item[4.] \begin{itemize}
              \item[t)] c\(_{t}\)(\idx{1}) :- \(\top_{h}\)( \idx{1}\typ{c} ).
              \item[h)] c\(_{h}\)(\idx{1}) :- c\(_{c}\)(\idx{1}), ( d\(_{h}\)(\idx{1}); e\(_{h}\)(\idx{1}) ).
              \item[c)] c\(_{c}\)( \typ{c} \und \avm{X & \idx{1}\avm{Z
                    & \idx{2}} \\ Y & \idx{1}} ) :- b\(_{t}\)(\idx{2}).
          \end{itemize}\vspace{1ex}
\item[5.] \begin{itemize}
              \item[t)] d\(_{t}\)(\idx{1}) :- \(\top_{h}\)( \idx{1}\typ{d} ).
              \item[h)] d\(_{h}\)(\idx{1}) :- d\(_{c}\)(\idx{1}).
              \item[c)] d\(_{c}\)( \typ{d} \und \avm{Y & \idx{1} \\ Z &
                  \idx{1}} ).
          \end{itemize}\vspace{1ex}
\item[6.] \begin{itemize}
              \item[t)] e\(_{t}\)(\idx{1}) :- \(\top_{h}\)( \idx{1}\typ{e} ).
              \item[h)] e\(_{h}\)(\idx{1}) :- e\(_{c}\)(\idx{1}).
              \item[c)] e\(_{c}\)(\typ{e}).
          \end{itemize}
\end{itemize}
}

Take for example the encoding for the type \typ{c}. The relation
c\(_{c}\) imposes the description language constraints for type
\typ{c} on the argument of the relation. Here the link between
relational level and the description language level is made. The
argument of the relation is assigned the correct type, and the
consequent of the type definition for \typ{c} (cf.\ group~1 theory in
figure~\ref{modelling-constraints}) is conjoined to that. No inherited
constraints are specified here, only the constraints directly imposed
on the type. Whenever in a group~1 theory a type restriction is
specified, in the corresponding group~2 theory, in the type's
c-relation -- where all description language restrictions are encoded
-- a call to the type's t-relation is used.  For example, the
specification \avm{X\|Z & b} in the type definition for type \typ{c}
is encoded in the relation c\(_{c}\) as a call to the relation
b\(_{t}\).  This ensures that all constraints on the type, including
the ones inherited, are applied whenever a type restriction is
specified.  Note that the appropriateness conditions for the argument
of a t-relation are preserved under this approach, since every
c-relation ensures that the argument is assigned the correct type.

The relation c\(_{h}\) references the constraints on type \typ{c}
which were collected in c\(_{c}\) and adds the constraints defined for
the subtypes of \typ{c} all the way down to the most specific
subtypes. In the example there is one more level to the most specific
types \typ{d} and \typ{e}. This step ensures that every object in the
denotation of the type \typ{c} also is in the denotation of one of the
most specific subtypes of \typ{c}, basically enforcing a limited
version of the closed world interpretation introduced in section
\ref{signature-section}. 

Finally, the relation c\(_{t}\) imposes on its argument all
constraints on type \typ{c}, those directly specified for the type,
and those inherited. This is done by collecting the constraints for
type \typ{c} from the constraint hierarchy all the way from the most
general type \(top\) down to a most specific sybtype of \typ{c}.

Summing up, this encoding technique provides us with a general
mechanism for transferring theories expressed using class 2
implicative statements into an architecture in which a theory has to
be formulated on the relational level. The ``look and feel'' of the
original formulation is preserved.  It is clear that one cannot expect
the linguist to re-encode a theory in such an awkward way, but such a
re-encoding could be done in a automated compilation step transferring
the grammar written by the linguist into machine friendly code.

In \hpsgII\ only very few types are actually constrained, i.e.\ appear
as the type of the antecedent of an implicative constraint. Also, most
currently implemented grammars are only used to query for (headed)
phrases.  The result is that only few relations need to be defined to
work with the system. Therefore most current implementations of HPSG
in a group~2 system do not follow anything like the general encoding
scheme introduced above.  On the other hand, an implementation close to
the original HPSG theory is only possible if constraints on any kind
of object can be expressed and queried separately from the top-level
signs. Only such an implementation can be produced and used by the
linguist to provide valuable feedback for a rigid and complete
formalization of the linguistic theory. Therefore a general automatic
translation procedure between the linguistic theory and the relational
encoding along the lines described above seems desirable. It allows
working close to the original linguistic theory, enables a modular
development and testing of linguistic theories, while at the same time
supplying runnable code.\footnote{The TFS encoding \citeN{Meurers93}
  (group~1) was transformed by hand (and sed) into a CUF grammar
  (group~2) along the lines described above.  The performance results
  in CUF were comparable to those of the original grammar running
  under TFS.}

\section{Modules of an HPSG theory: Special linguistic types of constraints 
  and choices for encoding them}
\label{ling-constraint-section}

\subsection{Specifying the lexicon} \label{lexicon} 

Like most current HPSG-style analyses, the theory proposed in HN is
strongly based on lexical specification.  In the light of the
discussion of the formal background of HPSG in the previous sections,
the set of constraints which linguists refer to as `the lexicon' is a
collection of implicative constraints no different from any other
constraint in the grammar (e.g.\ the principles).  Still there is a
reason why lexical information deserves a second look. In any normal
grammar the constraints making up the lexicon vastly outnumber the
rest of the constraints of the grammar.  From a linguistic as well as
a computational point of view it is therefore interesting to analyze
how generalizations over this big pool of constraints can be encoded.

The necessity to express lexical generalizations has long been
recognized. \citeN{Shieber86} comments the situation as follows:
``First, we can come up with general techniques for expressing lexical
generalizations so as to allow lexical entries to be written in a
compact notation. \ldots\ devices as templates and lexical rules in
PATR-II \ldots\ serve just this purpose: to express lexical
generalizations. As grammars rely more and more on complex lexical
encodings (as is the current trend in unification-based formalisms)
these techniques become increasingly important. Second, the formalism
can be extended in various ways to be more expressive''(p.\ 36). We
will see in this section that Shieber's first point directly carries
over to the current HPSG setup. The lexical rules are part of the HPSG
architecture, and the templates (now often called {\it macros\/}) are
used in implementations of HPSG grammars.  Regarding Shieber's second
point, we will show that the HPSG formalism is powerful enough to
allow for lexical generalizations to be expressed within the general
constraint formalism.

Basically, two kinds of generalizations over lexical information are
used in HPSG. Instead of completely specifying each lexical entry
separately, properties of whole classes of lexical entries are
specified. On the other hand, the existence of certain lexical entries
is related to the existence of other entries. After an introduction to
the basic setup of the lexicon, both kinds of generalizations and the
formal mechanisms for expressing them are discussed.

\subsubsection{Basics}

In the simplest case, a lexicon is a constraint on objects of type
\typ{word}.\footnote{In common HPSG theories \typ{word} is the only
  ``lexical'' type. In case one wants to postulate phrases in the
  lexicon (i.e.\ for proper names or bare plural noun phrases) a
  separate lexical type \typ{lex-phrase} could be introduced.  The
  type \typ{lex-phrase} would be defined as subtype of \typ{phrase},
  with sister-type \typ{struc-phrase}. The attribute DTRS should then
  be defined as appropriate for \typ{struc-phrase} instead of for
  \typ{phrase}. In the following, this issue is ignored and \typ{word}
  is assumed to be the only lexical type. Other kinds of lexical types
  will play a role though in the discussion of the so called lexical
  type hierarchies in section~\ref{lex-classes}.} Since the set of
possible lexical objects which are admitted by the linguistic theory
need to be constrained, in an HPSG grammar there is no way around a
statement defining the basic lexicon like that given in figure
\ref{lex}.

\myfig{lex}{The basic lexicon: a type definition for the lexical type
  \typ{word}}{\(\typ{word} \hspace{1em} \impl \hspace{1em} \left(L_1
  \; \oder \; \ldots \; \oder \; L_i \; \oder \; \ldots \; \oder \;
  L_n\right)\)}

In the terminology introduced in section~\ref{constraint-section}, the
figure~\ref{lex} shows a type definition for type \typ{word} with a
disjunctive description as consequent. Each \(L_i\) is a lexical
entry.  The effect of this implicative constraint is that every
lexical element has to be licensed by one of the disjuncts. There is
no way for a lexical object in a syntactic tree to ``sneak through''
unconstrained, which could be the case if other specifications were
included in the antecedent.

\subsubsection{Generalizing over a set of entries} \label{lex-classes}

The question we want to answer in this section is: How can the same
information be specified on several lexical entries? Referring to
figure~\ref{lex} this comes down to asking: How can information be
specified on several lexical entries \(L_i\) without repeating it in
the description of each disjunct. There are two possible answers to
this question and one theoretically less interesting solution which at
least avoids syntactic repetition. In the following discussion of the
three possibilities, we speak of a description \(C\) which we want to
specify on a set of lexical entries \(\cal{G}\).

\mysection{Expressing lexical generalizations outside of the lexical
  type definition}

For the first possibility, one starts with leaving the description
\(C\) out of the specification of each of the \(L_i\) disjuncts in
\(\cal{G}\).  The provisional result is a lexicon in which the lexical
entries in \(\cal{G}\) are too unrestricted.  Next, the constraint
\(C\) is introduced in the theory as an additional object definition.
The antecedent is specified to describe exactly the objects which are
described by the lexical entries in \({\cal G}\), i.e.\ it specifies a
property unique to the elements of \({\cal G}\).\footnote{If the
  elements of the set \(\cal{G}\) do not have a unique property to
  distinguish its elements from all other linguistic objects, such a
  unique property needs to be artificially introduced in each of the
  concerned \(L_i\) specifications in the \typ{word} type definition.
  However, such a specification naturally contradicts the original
  intention to specify information without repeating it in each
  disjunct. It is questionable whether the set of entries does form a
  natural set over which one should express generalizations.  The
  approach in such a case becomes equivalent to (and as theoretically
  uninteresting as) the macro approach described below.} The
consequent of the object definition is the constraint \(C\).

We have already seen an example for this encoding technique in section
\ref{group1-section}. The object definition of figure
\ref{obj-def-arg-raise} (p.~\pageref{obj-def-arg-raise}) was defined
to attach the argument raising specifications to auxiliaries.

\mysection{Lexical type hierarchies}
\label{lex-type-hierarchies}

The second possibility is to introduce subtypes of the lexical type
\typ{word} and attach the description \(C\) to the consequent of the
type definition of a subtype. Since under a closed world
interpretation of a type hierarchy every object that is of a type
\typ{t} also has to be described by one of its subtypes, the real work
under this approach is to make sure that via manipulation of the
signature we manage to have one of the subtypes of \typ{word} describe
exactly those \typ{word} objects which belong to the set \(\cal{G}\)
(every element of which we want to constrain with \(C\)).  Like in the
first encoding proposed above, this requires us to have a group
defining property for the elements of \(\cal{G}\).  Following the
method described on pp.~\pageref{type2encoding} of section
\ref{group1-section}, the description of the group defining property
then has to be encoded in a type, i.e.\ a subtype of \typ{word}.  An
example for such an encoding of a complex description in the signature
definition of a type is given in figure~\ref{type-def-aux-word} (pp.\ 
\pageref{type-def-aux-word}) where the type \typ{aux-word} is
introduced to reference all auxiliary-verb words.  Once the subtype of
\typ{word} is introduced in the signature, it can function as
antecedent of an implicative constraint and can specify the lexical
entries of \(\cal{G}\) as displayed in figure
\ref{lex-type-hierarchy-def}. Note the conjunct \(C\) which is the
description valid for all disjuncts, i.e.\ for all lexical entries
which are of type \typ{word-subtype}.

\myfig{lex-type-hierarchy-def}{Type definition for a lexical subtype}{
  \(\typ{word-subtype} \hspace{0.6em} \impl \hspace{0.6em} (L_j
  \;\oder\; \ldots \; \oder\; L_k) \;\und \; C\)}

When this approach is used to express several different
generalizations over different sets of lexical entries, the result is
a complex subtype hierarchy below \typ{word}, which has sometimes been
called a {\it lexical type hierarchy\/}.

Comparing this lexical type hierarchy approach to the first proposal,
one notes that the encoding of a complex description in a type (which
complicates the signature and restricts the antecedents to class 2
descriptions) is necessary because of theoretical considerations in
the lexical type hierarchy approach only. This is caused by the fact
that in the lexical type hierarchy approach a subtype of the lexical
type is introduced to get a hold on the lexical entries of the set
\(\cal{G}\). The correct denotation of this subtype has to be ensured
by manipulating the signature. In the first approach there is no
theoretic reason for restricting the implication antecedent to types.
However, in practice, the restriction of current computational systems
to type definitions forces the user in both approaches to encode the
class 2 antecedents describing the group defining property as types.

\mysection{Macros as syntactic placebo}
\label{placebo}

By now, the person interested in implementing a grammar is probably
rather desperate. Since, as already mentioned several times, current
computational systems do not allow the grammar writer to express
implications with complex antecedents, both proposals made so far for
expressing lexical generalizations have a fundamental flaw. They force
the implementer to encode additional information in the signature,
which severely complicates the hierarchy.\footnote{The reader thinking
  that ``severely complicates'' is an overstatement here, is reminded
  that encoding the relatively simple description of a word with a
  verbal head and \avm{AUX & plus} specification already demanded 19
  signature definitions (cf.\ figure~\ref{type-def-aux-word} on p.\ 
  \pageref{type-def-aux-word}).} In order to at least avoid writing
the same description over and over in the specification of lexical
entries, a syntactic grouping of information under one name is an
emergency solution that is better than nothing.  The idea borrowed
from the template mechanism of PATR-II is to give a name to a set of
descriptions and to use that name in place of the descriptions.  These
syntactic abbreviations nowadays are usually called {\it macros\/}.
The use of the macro mechanism for lexical specification is displayed
in figure~\ref{macro-def}.

\myfig{macro-def}{An example type definition for the lexical type
  \typ{word} using a macro \(M\)}{ \(\typ{word} \hspace{0.6em} \impl
  \hspace{0.6em} \left((L_1 \und M) \; \oder \; L_2 \; \oder \; (L_3
  \und M)\right)\)}

The macro name \(M\) abbreviates the description \(C\). It is
conjoined to each of the lexical entries belonging to the set
\(\cal{G}\). In the example of figure~\ref{macro-def}, \(L_1\) and
\(L_3\) belong to the set while \(L_2\) does not.

Before we turn to illustrating the macro mechanism, let us make a
remark on lexical specification in those computational architectures
which express grammatical constraints on the level of a relational
extension of the description language (group~2 architectures).  So
far, the discussion of the three possibilities for expressing
generalizations over lexical entries were all based on an architecture
like that of \hpsgII, in which the domain of objects can be directly
constrained (group~1 architectures).  Without using the technique for
expressing group~1 theories in group~2 architectures described in
section~\ref{group2-section}, the first two possibilities for
expressing generalizations discussed above are not available in a
group~2 setup since the domain cannot be directly constrained.  One
therefore only has the possibility to use the macro method, the
syntactic grouping of descriptions under a name. This also is the case
for the specification of lexical entries in those group~2 systems
which use a phrase structure backbone. As a result, the phrase
structure backbone based ALE and Troll implementation documented here
makes heavy use of macros to specify the lexicon.

To facilitate the presentation of examples from the implemented
grammar, we now briefly introduce the macro mechanism of
ALE.\footnote{Note that, since macros are abbreviations for (ALE)
  descriptions, they cannot contain calls to definite clauses.} Figure
\ref{unmarked} shows a simple definition of a macro in the ALE
implementation.

\myfig{unmarked}{Definition of macro ``unmarked''}
{{\tt unmarked macro (synsem:loc:cat:marking:unmarked).}}

A call to the macro, i.e.\ {\tt @unmarked}, is equivalent to writing
\avm{SYNSEM\|LOC\|CAT\|MARKING & \typ{unmarked}}, i.e.\ the macro
``unmarked'' denotes \typ{sign} objects whose marking values are of
type \typ{unmarked}. 

A macro can have parameters passed to it, allowing more versatile
definitions.

\myfig{head}{Definition of macro ``head''}
{{\tt head(X) macro (synsem:loc:cat:head:X).}}

Here a call, e.g.\ {\tt @head(verb)}, abbreviates the description of a
sign, whose head value has the value passed as the argument -- the
type \typ{verb}. 

Other macros can be called in the definition of a macro. A restriction
is that macro definitions are not allowed to be recursive. By using
the call to a macro as a specification in the definition of another
macro, hierarchies of definitions are formed.  These hierarchies are
subsumption hierarchies of descriptions. They have no theoretic status
in the architecture of HPSG and should not be confused with type
hierarchies as part of the signature. In the absence of other
theoretically more satisfying possibilities (i.e.\ in the current
implementation environments), such macro hierarchies are very useful
for hierarchically grouping information. In the implemented grammar,
the lexical entries were provided with the same information through
such a hierarchy of lexical macros. It is displayed in figure
\ref{lex-hierarchy}.

{\tabcolsep.2em
\renewcommand{\arraystretch}{2} 
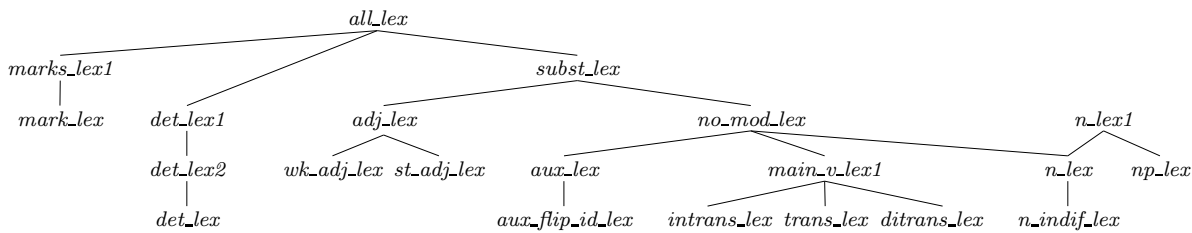
\myfig{lex-hierarchy}{The lexical macro hierarchy}{ \scriptsize\it
\begin{tabular}[t]{ccccccccccc}
\mcc{6}{\n{1}{all\_lex}}\\

\n{2}{marks\_lex1} & \mcc{7}{\n{3}{subst\_lex}}\\

\n{4}{mark\_lex} & \n{6}{det\_lex1}\hspace{2em} &
\mcc{2}{\n{7}{adj\_lex}} & \mcc{4}{\n{5}{no\_mod\_lex}} &
\mcc{2}{\n{n1}{n\_lex1}} \\ 

& \n{11}{det\_lex2}\hspace{2em} & \n{12}{wk\_adj\_lex} &
\n{13}{st\_adj\_lex} & \n{8}{aux\_lex} &
\mcc{3}{\n{9}{main\_v\_lex1}}& \n{10}{n\_lex} & \n{np}{np\_lex}\\ 

& \n{21}{det\_lex}\hspace{2em} & & & \n{14}{aux\_flip\_id\_lex} &
\n{15}{\hs intrans\_lex} & \n{16}{trans\_lex} &
\n{17}{ditrans\_lex\hs} & \n{ni}{n\_indif\_lex} & \\ 

\nc{1}{2} \nc{1}{3} \nc{1}{6} \nc{2}{4} \nc{3}{5} \nc{3}{7} \nc{5}{8}
\nc{5}{9} \nc{5}{10} \nc{6}{11} \nc{7}{12} \nc{7}{13} \nc{8}{14}
\nc{9}{15} \nc{9}{16} \nc{9}{17} \nc{n1}{np} \nc{n1}{10} \nc{10}{ni}
\nc{11}{21}
\end{tabular}
\vspace*{-5ex}}}

\renewcommand{\arraystretch}{1} 

The macros become more specific the further down in the hierarchy
one goes.  Note that contrary to type hierarchies, where we know that
every linguistic object has to be of a most specific type, any of the
macros in the hierarchy can be used to specify a lexical entry, not
just the most specific ones. 

Part of the information grouped in the lexical macro hierarchy above
serves to specify theory independent linguistic information (e.g.\ 
HEAD or CASE specification) and linguistic information driving a
specific analysis (e.g.\ argument raising, NPCOMP value), other
specifications seem to have a different status in supplying more
general HPSG architecture mechanisms with information (e.g.\ NONLOCAL,
QSTORE values). It seems obvious that this wealth of information and
its dependence structure should find its way into a proper HPSG device
-- even though this section has illustrated that in most architectures
(group~2 and those of group~1 which do not allow complex antecedents)
it is not clear what such a device should look like.

The macro hierarchies have often been confused with the lexical type
hierarchies discussed on p.~\pageref{lex-type-hierarchies}. Macro
hierarchies define abbreviations for complex descriptions by
hierarchically organized definitions of macros. The macros can then be
used in the specification of constraints. The confusion arises from
the fact that the lexical types are introduced as encodings of complex
descriptions (cf.\ p.~\pageref{lex-type-hierarchies}).  Nonetheless,
lexical {\it macro\/} hierarchies and lexical {\it type\/} hierarchies
are two completely different formal entities. The concepts and
terminology from one can therefore not be carried over to the other
without a redefinition.

Summing up, there are three distinct approaches for expressing
generalizations over lexical entries. It is possible to use several of
these approaches in one grammar, but one should be aware that they
employ different formal mechanisms to achieve their goal.

\subsubsection{Relating lexical entries}

\label{lex-rules}

Now that we have discussed basic lexical entries and the methods for
expressing generalizations over them, we can turn to the second type
of generalization used in HPSG-style theories: relations specifying a
lexical entry on the basis of another entry. Traditionally these
relations are called {\it lexical rules\/}.  Formally they are thought
of as ``meta-relations'', since they relate lexical entries, i.e.\ 
descriptions and not the objects which are related by ordinary
relations. Since a clear formalization of lexical rules as
meta-relations has not yet been given,\footnote{\citeN{Pollard93} is
  a sketch of such a formalization.} we will only sketch under what
idea of functionality linguists have used the lexical rule formalism
in the formulation of linguistic theories.  Following this
introduction of the lexical rule mechanism, we will investigate how
far we can get by expressing the generalizations captured in lexical
rules as meta-relations strictly on the level of the description
language. We then take a look at a computational mechanism which in
the ALE system provides some of the functionality of lexical rules. As
illustration of that computational mechanism, we show how the PVP
extraction lexical rule of HN can be encoded in ALE. The discussion of
lexical rules ends with some remarks on the (non-)correspondence
between linguistic and computational generalizations. Finally, as a
possible theoretical and computational alternative to the lexical
rules, an extension of the description language, the {\it named
  disjunction\/} mechanism, is introduced.

\mysection{Lexical rules}

In traditional terms, a lexical rule is a pair of two
meta-descriptions, one for each class of descriptions related. Views
vary significantly regarding the interpretation of the information on
the two meta-descriptions and that of the lexical rule as such.

\mysubsection{Application criterion:}\nopagebreak Two different
assumptions are made regarding the question of when a lexical rule
applies to a lexical entry (a description): Either the lexical entry
has to be {\it more specific than\/} the left-hand side of the lexical
rule, or the lexical entry has to be {\it consistent with\/} the
left-hand side of the lexical rule. Using the ``more specific than''
option is the more restrictive approach, which corresponds to the view
of information in the lexical entry triggering the lexical rule.  The
``consistent with'' option on the other hand additionally allows
descriptions which are part of the left-hand side of a lexical rule to
further narrow down the denotation of the lexical entry, which makes
it impossible to distinguish between the specifications of the entry
and that of the rule.

Often the term {\it subsumption\/} is used for the ``more specific
than'' relation and {\it unifies with\/} instead of ``is consistent
with''.  One should be aware that using these terms when talking about
the lexical rule mechanism is very informal. Unification and
subsumption are not even part of the description language.\footnote{In
  the Carpenter setup unification and subsumption play a role on the
  semantic level, not on the syntactic level of the description
  language. In the King setup, on which \hpsgII\ can be based,
  unification and subsumption play no role at all. Cf.~\ref{ontology-section}} It is therefore not possible to ``lift'' the
two relations from the description language level to the meta
description level.  In fact, the issue of lexical rules as
meta-descriptions is even more complicated, since even a formally
defined ``meta-subsumption'' or ``meta-unifies-with'' would be
inappropriate because entities on two different language levels need
to be related.  A lexical entry is a {\it description\/} whereas a
lexical rule is a pair of {\it meta-descriptions\/}. The easiest
solution would be to drop the idea of lexical rules as meta-relations
and use ordinary relations on objects to capture the desired
functionality. However, the following section will show that such an
encoding misses some of the functionality associated with lexical
rules. Before we analyze this matter further, let us finish the
description of the functionality associated with lexical rules in
intuitive terms.

\mysubsection{The nature of the transfer:}\nopagebreak Once the rule
applies, the tags specified in the left-hand side ``pick out''
information from the lexical entry and ``supply'' it to the
occurrences of the tags in the right-hand side.  This information
transfer can be done via ``{\it unification\/}'' or {\it
  copying\/}.\footnote{The tags which are often used to notate the
  information transfer in lexical rules as meta-descriptions should
  not be confused with the tags used in the description language. A
  description language tag is interpreted as identity of the set of
  objects denoted, whereas the lexical rule ``meta-tags'' specify
  identity of descriptions.} The specifications on the left-hand side
of a lexical rule therefore has two functions: it triggers rule
application and provides information handles.

\mysubsection{Unchanged specification carryover:}\nopagebreak
Information ``not changed'' by the lexical rule is assumed to be
transferred to the output. Several possibilities arise here concerning
the interpretation of ``not changed'' and again the nature of the
transfer.

A formalization of lexical rules as meta-descriptions will have to
refer to these choices.  One should bear in mind that the exact
formulation chosen has immediate consequences on the linguistic
theories which can be formulated using the lexical rule mechanism.
\citeN{Hoehle94} shows that a lexical rule mechanism with
unification as the application criterion does not work properly in a
grammar where complement extraction is done via lexical rules and
argument raising \`a la Hinrichs and Nakazawa is used to construct
the verbal complex.

Looking at the theory proposed in HN a further desideratum of the
lexical rule mechanism presents itself.  The PVP extraction lexical
rule and the split-NP lexical rule share a common mechanism.  Roughly
speaking, partially extracted constituents are related to remnants in
base position. This is implicit in the paper, but cannot be expressed
in the formulation of the lexical rules. Using a hierarchically
structured tool to relate lexical entries, it should be possible to
express different levels of generalizations and capture the
resemblance. To our knowledge, no formulation of such hierarchical
lexical rules is proposed in the literature.

\mysection{Lexical rules as part of the description language}

In this section we investigate how far we can get by using only the
description language to capture the functionality of lexical rules.
Figure~\ref{full-lex} shows a type definition for \typ{word}, which
defines an extended lexicon including lexical rules.\footnote{Several
  equivalent formulations are possible. For example, two subtypes of
  \typ{word} can be introduced, one for the standard lexicon
  (\typ{normal\_word}) and one for the results of lexical rules
  (\typ{lex\_rule\_word}).  The type \typ{lex\_rule\_word} has an
  appropriate, word typed IN attribute.  The ``out'' specifications
  are specified directly on the \typ{lex\_rule\_word} object. This
  encoding is basically the one used in the TFS grammars documented in
  \citeN{Kuhn93} and \citeN{Keller93}.}

\newcommand{\storeAvm}[1]{\idx{#1}\avm{STORE & \XlstI{\lrDefOut{1}}}}
\newcommand{\lrDef}[1]{\hspace{-0.78em}\avmt{lex\_rule}{IN & \(D_#1\)
    \\ OUT & \(E_#1\)}}
\newcommand{\lrDefOut}[1]{\hspace{-0.78em}\avmt{lex\_rule}{OUT &
    \idx{#1}}}

\myfig{full-lex}{A lexicon with added lexical rule
  relations}{\(\typ{word} \hspace{1em} \impl \hspace{1em} (L_1 \;
  \oder \; \ldots \; \oder \; L_i \; \oder \; \ldots \; \oder \; L_n
  \; \oder \;\;\storeAvm{1})\)}

The formulation makes use of the junk slot technique discussed on
p.~\pageref{relationsIntoDescriptions} of section~\ref{group1-section}
for encoding relations on the level of the description language. The
type \typ{word} is assumed to have an additional appropriate attribute
STORE which is \typ{list}-valued. Furthermore a new type
\typ{lex\_rule} is introduced below \(\top\), having IN and OUT as
appropriate attributes with \typ{word} values. The type definition for
\typ{lex\_rule}, in which the different lexical rules are specified,
is given in figure~\ref{lex-rule-type-def}.

\myfig{lex-rule-type-def}{Specifying the lexical
  rules}{\typ{lex\_rule} \hspace{1em} \impl
  \hspace{1em} (\lrDef{1} \oder \hspace{0.2em} \ldots \oder
  \hspace{0.2em} \lrDef{m})}

How does this description language encoding work? The definition of
the lexicon in figure~\ref{full-lex} is a type definition on the type
\typ{word} just like the basic lexicon of figure~\ref{lex}. Every
object of type \typ{word} has to satisfy one of the standard lexical
entries \(L_i\) or it satisfies a special lexical entry \idx{1}. The
special entry \idx{1} is the description which is the OUT value of the
\typ{lexical\_rule} object residing in the special entry's junk slot
attribute STORE.\footnote{Note that the description of the special
  entry is a cyclic structure. This can be avoided if one instead
  introduces a \typ{lex\_rule} subtype of word as described in the
  previous footnote. However, the encoding with the cyclic structure
  has the advantage of clearly separating lexical rule objects from
  words. This makes it easier to compare the lexical rules in the
  description language setup with the other formalizations of lexical
  rule relations.}

The admissible \typ{lex\_rule} objects are defined in figure
\ref{lex-rule-type-def}. Because of the special lexical entry \idx{1}
in the type definition of word, every OUT description of a lexical
rule object is a possible word. To relate the OUT value to another
lexical entry, structure sharing between the descriptions \(D_i\) and
\(E_i\) of each disjunct in figure~\ref{lex-rule-type-def} needs to be
specified. Note that the appropriateness conditions for
\typ{lex\_rule} ensure that the value of the IN attribute is of type
\typ{word}. It therefore has to satisfy the type definition for word,
i.e.\ one of the lexical entries of figure~\ref{full-lex}. Naturally
that lexical entry can again be the output of a lexical rule.

Let us illustrate the definition of a lexical rule, i.e.\ one of the
disjuncts of figure~\ref{lex-rule-type-def}, with an example based on
the \hpsgII\ appendix signature.  For expository purpose only, assume
we want to write a lexical rule, which extracts the subject of
intransitive verbs with base verb-form and inserts it on the inherited
slash.  The resulting encoding is shown in figure
\ref{lex-rule-encoding}.

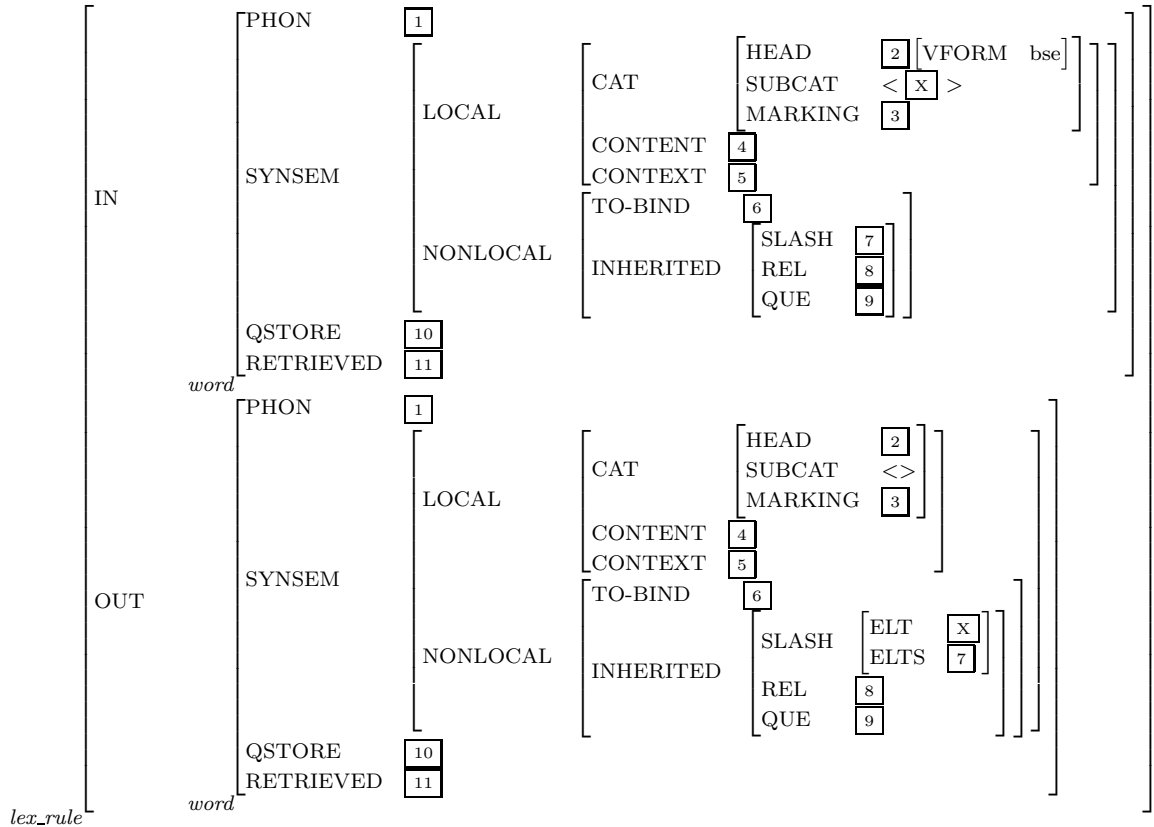
\myfig{lex-rule-encoding}{An example for a lexical rule description}{
  \avmt{lex\_rule}{IN & \avmt{word}{PHON & \idx{1}\\SYNSEM &
      \avm{LOCAL & \avm{CAT & \avm{HEAD & \idx{2}\avm{VFORM & bse} \\ 
            SUBCAT & \XlstI{\idx{X}}\\MARKING & \idx{3}}\\CONTENT &
          \idx{4}\\CONTEXT & \idx{5}}\\NONLOCAL & \avm{TO-BIND &
          \idx{6}\\INHERITED &
          \avm{SLASH&\idx{7}\\REL&\idx{8}\\QUE&\idx{9}}}}\\QSTORE &
      \idx{10}\\RETRIEVED & \idx{11}}\\ OUT & \avmt{word}{PHON &
      \idx{1}\\SYNSEM & \avm{LOCAL & \avm{CAT & \avm{HEAD & \idx{2} \\ 
            SUBCAT & \Xelst\\MARKING & \idx{3}}\\CONTENT &
          \idx{4}\\CONTEXT & \idx{5}}\\NONLOCAL & \avm{TO-BIND &
          \idx{6}\\INHERITED & \avm{SLASH&\avm{ELT & \idx{X}\\ELTS &
              \idx{7}}\\REL&\idx{8}\\QUE&\idx{9}}}}\\QSTORE &
      \idx{10}\\RETRIEVED & \idx{11}}}}

The description of the verb-form (tagged \idx{2}) on the input ensures
that only \typ{bse} verbs can undergo this lexical rule. The tag \idx{X}
references the (only) subcat list element of the input and is
identified with an additional element of the slash set of the output
description.  The tags \idx{1} through \idx{11} ensure that all other
specifications are structure shared between the input and the output.

What effect does this encoding have on the choices mentioned above for
lexical rules as meta-relations: the application criterion, the nature
of the transfer, and the carrying over of unchanged information?

To make the following discussion easier to read, we call one of the
disjuncts of figure~\ref{lex-rule-type-def}, a description of
admissible lex-rule objects, \(R\). It's IN attribute bears the
specification \(D\) and its OUT attribute the description \(E\).
Furthermore \(L\) is a lexical entry \(L_i\) of figure~\ref{full-lex}.

When does the lexical rule \(R\) apply to a lexical entry \(L\)?  The
input specification \(D\) of the lexical rule \(R\) describes a set of
word objects.  Only those word objects which also satisfy the type
definition for \typ{word} satisfy our theory. Looking at that type
definition in figure~\ref{full-lex}, this means that the word objects
which are denoted by \(D\) also have to be in the denotation of a
lexical entry \(L\) to satisfy the theory. The result is that a
lexical rule applies to those objects which are in the intersection of
the denotation of the lexical entry \(L\) and the input description
\(D\).  This semantic characterization on the syntactic side
corresponds to the conjunction of the descriptions of \(D\) and \(L\),
which in the Troll setup is computed as the unification of the feature
structure normal form representation of both descriptions. In the
informal terminology of the lexical rule as meta-relation sketch
above, the setup of lexical rules in figure~\ref{full-lex} therefore
entails that a lexical rule applies if its input description ``unifies
with'' the description of a lexical entry.

Now to the nature of the transfer. To state that two descriptions are
partly identical, structure sharing (token identity) between parts of
the descriptions of the IN attribute's value and that of the OUT
attribute needs to be specified.  Using the expressive power of the
logic only, it is not possible to demand type identity without
explicitly using the same description twice.  The logic does not
contain anything corresponding to the intuitive notion of copying.
The result for our lexical-rule-like mechanism formulated within the
description language is that the only way to relate the input and the
output descriptions \(D\) and \(E\) is by specifying structure sharing
between their sub-descriptions.

Similarly, it is not possible to specify any ``carrying over'' of
``unchanged'' specifications. All ``carrying over'' is done by
explicitly specifying structure sharing, i.e.\ all descriptions have
to be explicitly specified on the OUT attribute. The grammar writer
therefore has to decide for each lexical rule which information is to
be preserved and ensure that it gets transferred from the input to the
output of the rule. For two reasons it is sometimes very difficult to
fulfill this task. Assume the following signature:

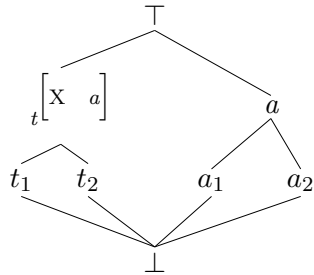
\myfig{ex-signature}{An example signature}{
\begin{tabular}[t]{ccccc}
& & \n{top}{\(\top\)} & &\\[3ex]
\mcc{2}{\n{t}{\avmt{t}{X & \typ{a}}}} & & \mcc{2}{\n{a}{\typ{a}}}\\[3ex]
\n{t1}{\(t_1\)} & \n{t2}{\(t_2\)} & & \n{a1}{\(a_1\)} &\n{a2}{\(a_2\)}\\[4ex]
& & \n{bot}{\(\bot\)} & &
\nc{top}{t} \nc{top}{a} \nc{a}{a1} \nc{a}{a2} \nc{t}{t1} \nc{t}{t2}
\nc{t1}{bot} \nc{t2}{bot} \nc{a1}{bot} \nc{a2}{bot}
\end{tabular}}

Now take the description \avmt{\(t_1\)}{X & \typ{\(a_1\)}} and suppose
we want to specify a second description so that it specifies all the
information of the first, except that the value of its X attribute is
specified as \(a_2\). This is only possible if we can access the type
of an object independent of its attribute specifications. No
computational system currently provides this possibility. The type
information \(t_1\) of the first description therefore can not be
specified on the second feature structure without copying the whole
\typ{t} object including the \avm{X & \(a_1\)} specification. The only
solution is to split the lexical rule into many different lexical
rules, one for each of the possible types of the objects which cannot
be carried over completely.  In the example above this results in two
lexical rules: one for \(t_1\) objects and one for \(t_2\) objects.
In every more complex case this leads to a serious multiplication of
lexical rules which loses some, possibly all originally intended
generality.

As a slightly more complex example based on the signature definitions
given in the appendix of \hpsgII\ suppose (for expository purpose
only) we want to specify a lexical rule, which relates all
non-predicative entries to predicative ones. The only thing the
lexical rule does is change the PRD attribute's value from \typ{minus}
to \typ{plus}. All signs having a PRD attribute (i.e.\ the signs with
substantive heads) can undergo this lexical rule. Regarding the
question of which specifications of the input are supposed to be
carried over to the output, the answer has to be that everything
except for the HEAD value should be passed, since the HEAD value bears
the PRD specification. The unintended but unavoidable result is that
the type of the HEAD value object of the lexical entry (i.e.\ 
\typ{prep}, \typ{noun}, \typ{verb}, or \typ{adj}) is lost after the
application of the lexical rule. The only solution is to split the
original lexical rule into one for each HEAD subtype.  These lexical
rules then carry the specification of the HEAD subtype on the input
and the output.  Note that this completely loses the originally
intended generalization over all non-predicative signs.

The second problem regarding the specification transfer from input to
output is caused by the fact that the feature geometry of lexical
entries which can undergo a single lexical rule can vary
significantly. In some cases the grammar writer therefore has to split
up a lexical rule and specify which information gets carried over
depending on the shape of the input. For example for the
prd-lexical-rule of the last paragraph, a separate lexical rule would
have to be made for verbs (to ensure the VFORM specification is
carried over), nouns (to ensure the same for CASE), prepositions
(PFORM), and for the only remaining substantial signs without specific
head attributes, the adjectives. Again, the generalization over all
substantial signs is lost.

Summing up, it is possible to encode some of the functionality
commonly associated with lexical rules on the level of the description
language. However, the resulting description language mechanism
commits the user to explicitly specify which information of the input
is supposed to be present on the output. We have seen that this can
cause problems. Nonetheless, the kind of description language
mechanism defined currently is the only way to include at least the
core of the functionality of lexical rules in a well-defined and
formal way inside of the feature logic on which \hpsgII\ is based.
The restriction of the description language mechanism to an explicit
transfer of specification is mainly caused by the problem to formalize
the notion of a transfer of ``unchanged specifications''. We will see
in the next section that the computational mechanism to encode lexical
rules used in the ALE system has exactly the same deficits as our
description language encoding.

\mysection{The PVP extraction lexical rule and its implementation in
  ALE}

\label{ALE-lex-rules}

Before turning to the specific ALE mechanism for lexical rules let us
comment on the procedural metaphors commonly used with lexical rules.
Even though lexical rules formally are binary relations without
explicit order, one usually speaks of an ``input'' description and an
``output'' description of a lexical rule.  The meaning of this is that
the lexicon without the lexical rules is usually set up so that it
contains certain lexical entries matching the input descriptions of
the lexical rules, but none which match the output descriptions. When
the procedural metaphor behind this terminology is transferred into a
procedural interpretation in the sense that lexical rules produce new
lexical entries (like in the ALE system), this can lead to unintended
results. The problem is that different application orders can produce
several instances of the identical lexical entry.

Take for example two lexical rules \(R_1\) and \(R_2\) which both
apply to a lexical entry \(A\).  Additionally assume \(R_1\) and
\(R_2\) modify different parts of \(A\) and that the intersection of
the denotation of the output description of either of the rules and
that of the input description of the other rule is not empty. In this
situation, a procedural interpretation of lexical rule application
would produce four new entries, one resulting from the application of
\(R_1\), one from the application of \(R_2\), one from applying
\(R_1\) and then \(R_2\), and finally one where first \(R_2\) and then
\(R_1\) was applied.  The last two possibilities cause two identical
lexical entries to be added to the lexicon.  Clearly such a production
of identical entries is an unwanted effect of the procedural
interpretation of lexical rules.

In the ALE system, lexical rules are basically a special notation for
unary phrase structure rules with an additional mechanism to alter the
morphology. When the grammar is compiled, the closure under lexical
rule application is computed. New lexical entries are produced and
added to the lexicon.  As mentioned above this can lead to identical
entries as the result of several possible orders of application of one
lexical rule on the output of another. Application is done with
unification as test, and the new entry is created with copies of the
information specified in the tags. Just as with the description
language formalization just discussed, only the information explicitly
specified on the right-hand side of the lexical rule finds its way
into the new entry.  The resulting problems noted above therefore
apply to the ALE formalism as well.\footnote{In the implemented
  grammar it was possible to avoid the problem regarding the different
  feature geometries of input and output lexical entries because of
  the homogeneous nature of the application domain of the lexical
  rules in the fragment (mostly verbal entries).} An additional
complication of the specification transfer issue can arise from the
procedural interpretation of the lexical rules.  Because of different
possible application orders, an insufficient specification of
propagation of information from input to output can lead to unwanted
unilateral subsumption between entries (in addition to identical
entries resulting from applying the same lexical rules in different
order).  Whether all unilateral subsumption can be eliminated by
changes to the lexical rules depends on the possibility to completely
pass unchanged information.  As discussed above this can sometimes
only be achieved at the loss of generality by splitting a lexical rule
into many different lexical rules.\footnote{In the implementation,
  extra PROLOG routines were used to detect and eliminate mutually and
  unilaterally subsuming lexical entries. With additional
  specifications on the lexical rules and entries it was possible to
  eliminate the unwanted lexical entries in the special case of the
  implemented grammar.}

To illustrate a lexical rule encoding in the ALE system, let us take a
closer look at the PVP extraction lexical rule which in HN drives the
analysis of possibly partial verb phrase extraction.  Figure
\ref{hn-pvp} shows the lexical rule as given in HN\footnote{The
  ordering on the comps list is inverted compared to the figure given
  in HN. Cf.\ section~\ref{valence-encoding} for a discussion of this
  issue. Also, the \avm{VFORM & \typ{bse}} specification on the input is
  only implicit in the formulation of the PVP extraction lexical rule
  given in figure 19 of HN.}.

\newcommand{\compsEl}[0]{\myavm{SYNSEM\|LOC & \myavm{CAT & \myavm{HEAD &
        \avmt{verb}{VFORM & \idx{1}} \\ VAL & \myavm{COMPS & \idx{2} \\ 
          SUBJ & \idx{3}}} \\ CONT & \idx{4}}} }

\newcommand{\lhsPVP}[0]{\myavm{SYNSEM & \avmNo{LOC\|CAT \myavm{HEAD
        & \avmt{verb}{VFORM & \typ{bse} \\ AUX & \typ{plus}} \\ VAL\|COMPS
        & \lst{\compsEl}{\idx{2}}} \\ NONLOC\|INHER\|SLASH \{\}}}}

\newcommand{\listDescHN}[0]{\typ{list}\((\neg\myavm{SYNSEM\|LOC\|CAT\|HEAD
    & \typ{verb}})\)}

\newcommand{\listDesc}[0]{\typ{list}\((\myavm{SYNSEM\|LOC\|CAT\|HEAD
    & \typ{noun}})\)}

\newcommand{\slashSetEl}[0]{\myavm{SYNSEM & \avmNo{LOC  \myavm{CAT & \myavm{HEAD &
          \avmt{verb}{VFORM & \idx{1}} \\ VAL & \myavm{COMPS & \elst \\ 
            SUBJ & \idx{3}}} \\ CONT & \idx{4}} \\ 
      NONLOC\|INHER\|SLASH \idx{5}}}}

\newcommand{\rhsPVP}[0]{\myavm{SYNSEM & \myavm{LOC\|CAT\|VAL\|COMPS &
      \listDescHN \\ NONLOC\|INHER\|SLASH & \(\left\{\slashSetEl\right\}\)}}}

\vspace*{-2ex}
\myfig{hn-pvp}{The PVP extraction lexical
  rule} {\hspace*{-0.8cm} \lhsPVP \\[2ex] {\Large \impl} \\[2ex]\hspace*{-0.8cm} \rhsPVP}

\vspace*{1ex}
Figure~\ref{pvp} shows the corresponding ALE encoding:
\vspace*{-1.6ex}
\myfigRef{pvp}
\begin{verbatim}
pvpelr lex_rule
   (In, @verb, @bse,
               @plus_aux,
        @val_comps([ (@verb, @vform(Vform),
                      @val_comps(Comps),
                      @val_subj(Subj),
                      @cont(Cont))
                    | Comps ]),
        @no_in_slash)

**>
   (Out, @val_comps( (L, list_n_sign) ),
         @in_slash([ (@verb, @vform(Vform),
                      @comps_sat,
                      @val_subj(Subj),
                      @cont(Cont),
                      @in_slash(L)) ]))
if
   pass2(In,Out)
morphs
   X becomes X.
\end{verbatim}
\vspace*{-2ex}\myfigTitle{The encoding of the PVP extraction lexical rule}

\label{list-np-type-mention}
Due to the systematic use of macros, the implementation encoding
closely resembles the lexical rule given in HN. The only difference
concerns the value of the COMPS attribute in the output description.
Since ALE does not support polymorphic types or negation of
descriptions, the ``list of non-verbal signs''-description on the
COMPS attribute cannot be expressed directly.  In the current theory,
restricting the COMPS value of the output of the PVP extraction
lexical rule to be a list of nominal signs, will also makes the
intended correct predictions.\footnote{The same reasoning holds for
  the constraint on the elements which can undergo argument raising on
  the comps list of auxiliary verbs. Just as in the PVP extraction
  lexical rule case, the grammar implemented restricts them to be
  nominal signs.} The negated descriptions therefore can be replaced
with the description of a list of nominal signs.  Due to the absence
of polymorphic types, the method for encoding descriptions in the
signature definition of a type (cf.\ section~\ref{group1-section},
pp.~\pageref{type2encoding}) was used to introduce a new type
\typ{list\_n\_sign} for ``list of signs with a nominal head''. This
type constraint was then used to restrict the value of the COMPS
attribute in the output of the lexical rule in figure~\ref{pvp}.

The call to the definite clause {\it pass2\/} at the end of the
lexical rule encoding of figure~\ref{pvp} takes care of the carrying
over of information not changed by the lexical rule. The definition of
the definite clause and the macro involved in passing all but the
changed COMPS and INHERITED information is given in figure
\ref{pvp-extras}.

\myfigRef{pvp-extras}
\begin{verbatim}
pass2(@core2(Qstore,Retr,Head,Marking,Npcomp,Val_subj,Val_spr,Cont,Conx,To_bind),
      @core2(Qstore,Retr,Head,Marking,Npcomp,Val_subj,Val_spr,Cont,Conx,To_bind))
   if true.

core2(Qstore,Retr,Head,Marking,Npcomp,Val_subj,Val_spr,Cont,Conx,To_bind) macro
   (word, (qstore: Qstore),
          (retr: Retr),
          @head(Head),
          @marking(Marking),
          @npcomp(Npcomp),
          @val_subj(Val_subj),
          @val_spr(Val_spr),
          @cont(Cont),
          @conx(Conx),
          @to_bind(To_bind)).\end{verbatim}
\vspace*{-2ex}\myfigTitle{The mechanism for passing the unchanged specifications
  (everything but comps and inherited slash)}

The parameter list of the macro core2 is used as an interface to all
relevant attribute specifications. In the definite clause pass2 the
macro core2 is called twice, once for the input and once for the
output. For both calls the same variables are passed as arguments,
ensuring identity between the relevant attribute specifications of
input and output.

Besides the PVP lexical rule driving the analysis of partial VP
topicalization, several others lexical rules were included in the
implementation to obtain a simple and consistent grammar: A
morphological rule relating the negative indefinite ``kein'' to the
various forms of ``ein''; several morphological rules with syntactic
effect obtaining finite, past participle, and substitute infinitive
forms of verbs; syntactic rules introducing unbounded dependencies in
lexicalized style for complement and subject extraction; and finally a
rule freeing the complement order of ditransitive verbs.

\mysection{Linguistic generalizations and computational compactness}

In the last sections we have been concerned with mechanisms for
expressing generalizations over the lexicon by relating lexical
entries. We sketched the functionality desired by the linguists,
showed that a formal mechanism inside of the logical setup of \hpsgII\ 
does not capture all of these desiderata, and displayed the
computational mechanism used in ALE which is equivalent to the
description language mechanism regarding the functionality captured.
The discussion so far was focussed on how the linguistic
generalization are expressed, i.e.\ on the functionality of the
mechanism.

Regarding computational mechanisms for expressing lexical rules, an
issue besides the functionality offered deserves some discussion. From
the computational point of view it is important whether the methods
used to express the linguistic generalizations have correspondents on
the computational side. In other words, in an ideal computational
setup, the methods used to express linguistic generalizations should
have a counterpart in a compact and elegant computational encoding
which can be dealt with using efficient algorithms. We have seen that
in the ALE system there is no such correspondence. The lexical rules
are only used to produce new lexical entries for a bigger lexicon,
which is then used in processing. This is particularly inelegant since
the new entries produced usually only differ in very few
specifications, namely those altered by the lexical rules.

There seem to be two possibilities to improve on this situation. One
could use the ALE mechanism with an alternative processing strategy.
It would supply new lexical entries whenever a lexical entry is needed
in the local tree being processed.\footnote{The same effect follows
  from using the description language mechanism introduced above in a
  group~1 setup.} However, such a straightforward application of
lexical rules ``on the fly'' is very inefficient since every time
something needs to be licensed by a lexical entry all lexical rules
have to be applied in all possible orders to collect all possibly
matching entries. A more complicated scheme could use some kind of
tabelling method to keep track of the lexical entries already
produced. But since it is completely unconstrained which
specifications can be changed by a lexical rule, in the general case
such a tabelling method will result in compiling out the lexicon upon
the first lexical lookup to make sure that all possible solutions are
found. Naturally this is even less desirable than compiling out the
lexicon beforehand.

The other possibility uses the insight that the lexical entries
resulting from lexical rule application usually only differ in very
few specifications compared to the number of specifications in the
lexical entry as a whole. Instead of expressing relations between
lexical entries -- which in a formal mechanism causes significant
problems regarding the transfer of unchanged specifications from input
to output -- one can use a single lexical entry and encode the
possible variations in that entry. Note that the variations cannot be
captured with simple disjunctions since the value of one specification
might depend on the value of another. Let us therefore call them
systematic covariations. 

\mysection{Using relations to encode systematic covariation in lexical entries}

The question we are interested in is how the lexical rules proposed by
the linguists can be (automatically) translated into a system of
relational dependencies introducing systematic covariation within
lexical entries.  A full discussion of this topic is beyond the scope
of this paper. We will therefore only sketch an approach using the
relational extension of the description language. As an alternative,
we will briefly introduce an extension of the description language
with so called ``named disjunctions'' in the next section.

The lexical rules and the mechanism for introducing systematic
covariation are intended to capture the same generalizations. Still,
they are very different in the way they work: The systematic
covariation encoded as relational dependencies in the lexical entries
capture the difference between a ``basic'' lexical entry and all those
lexical entries which are the result of (successively) applying the
lexical rules to that basic entry. Or in more computational terms: in
the covariation approach, calls to definite clauses are attached to
lexical entries. The definite clauses introduce disjunctive
specifications in a lexical entry. 

Lexical rules on the other hand relate lexical entries.  Compared to
the covariation setup they lift the relational dependencies from
within a lexical entry up to the level of the lexical entries. This
allows us to generalize over whole classes of lexical entries.  These
kind of generalizations over classes of lexical entries seem to be
lost in the covariation approach, since the covariation is encoded
inside of each lexical entry. This is not true though if we can use a
call to the same relation in all or a natural class of lexical entries
to introduce the intended covariation. To be able to define such a
relation, we need to know which lexical rules can apply in what order
on which lexical entries. This information can be retrieved from the
input and output specification of the lexical rules. By checking which
input of a lexical rule is compatible with which output, we can obtain
a graph representation of all possible application orders of lexical
rules to an entry. Figure~\ref{fsa} is such a graph for the lexical
rules which in the implemented grammar apply to the class of verbal
entries.

\parbox[t]{\textwidth}{\hspace*{0.3cm}\myfigRef{fsa}\psfig{figure=automat.idraw}
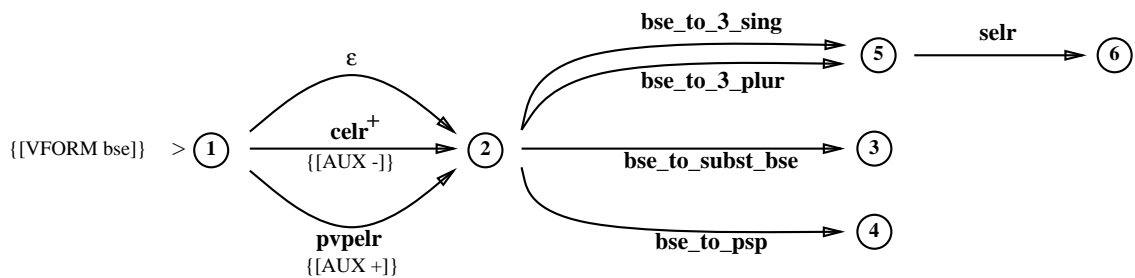
  \myfigTitle{The lexical rules applying to verbal elements and their
    interdependence}}

Each edge corresponds to the application of a lexical
rule\footnote{The abbreviated lexical rule names have the following
  expansions: subject extraction (SELR), partial vp extraction
  (PVPELR), finitivization (bse\_to\_3\_sing, bse\_to\_3\_plur),
  finitivization (finit.), and past participle formation
  (bse\_to\_psp, bse\_to\_subst\_bse). The celr\(^+\) is the
  complement extraction lexical rule modified in such a way that it
  allows the extraction of any number of complements greater than 0.
  It for example only has to be applied once to extract both objects
  of a ditransitive verb.  The $\epsilon$ stands for no change.}. Each
node relates to one solution of a definite clause attached to a
lexical rule. The specifications in curly brackets describe the set of
lexical entries for which the graph as a whole or a certain edge is
applicable. 

The graph displayed applies to verbal entries with \typ{bse}
verb-form, which in fact are the only verbal entries specified in the
basic lexicon of the fragment. The PVPELR only apply to auxiliary
verbs and the CELR only to full verbs. The restrictions which are
already ensured by the specification of the output of a lexical rule
do not need to be listed separately. For example the SELR only applies
to finite verbs. Since the finitivization rules leading to node 5
ensure that their result is finite, no additional specifications are
necessary. 

For the issue we are pursuing, the graph can be used as backbone for
an encoding of the lexical rule regularities in the relations called
in the lexical entries. Such a backbone is constructed by encoding the
graph as finite state automaton in one definite clause which is
attached to every lexical entry. Every node is a solution, i.e.\ a
possible lexical entry, and every label of an arc is a call to a
relation encoding the variation introduced by the corresponding
lexical rule.

In one of the Troll versions of the grammar Guido Minnen and I encoded
lexical rules in the way described. This drastically reduced the
number of lexical entries compared to the original compiled out ALE
lexicon. The result is a lexicon which captures at least some of the
linguistic generalizations behind the lexical rules in a
computationally usable way.

\mysection{An alternative mechanism: named disjunction}

\label{named-disjunction-page}

Before we leave the area of generalizations over lexical information,
let us briefly introduce an alternative mechanism using an extension
of the description language for expressing systematic covariation in
lexical entries: named disjunctions of complex descriptions used in
the specification of lexical entries.

To illustrate this mechanism we will look at an example from the
interaction of morphology and syntax. In HN (adapting a proposal from
\citeN{PollardHNM}) base verbs differ from finite verbs in three
respects.  The subject of a finite verb is encoded together with its
complements, whereas that of a base entry is specified in the separate
attribute SUBJ.  Nominative case as well as number and person
information is specified for the subject of finite verbs, but not for
that of base ones.  Finally the verb form is encoded in the attribute
VFORM. A fully expanded lexicon would contain seven unrelated entries
and the relation between these would be lost. Note also that the
described complex interaction of morphological and syntactic features
of finitivization cannot be encoded using simple disjunctions, since
the case and agreement assignment, as well as the valence encoding
depends on the verb form.  Figure~\ref{HNM} shows the resulting named
disjunction encoding.

\myfig{HNM}{A named disjunction expressing finitivization \`a la
  \citeN{PollardHNM} (\typ{cat} value shown)}
{\hspace*{-1.4cm}\avmt{cat}{HEAD & \avmt{verb}{VFORM & \nd{v}{\fin
        \dis \typ{bse}}} \\ VAL & \avm{SUBJ & \nd{v}{<> \dis <\idx{1}>}
        \\ COMPS & \nd{v}{< \left(\idx{1} \und \avm{\path{CASE} &
            \typ{nom}} \right) \;.\; \idx{2}> \dis \hspace{0.6em}
          \idx{2}} \\ SUBCAT & \(<\idx{1} \;.\; \idx{2}>\)}}}

The notation needs some explanation. The \path{F} abbreviates the path
starting with SYNSEM down to the attribute F. The angle brackets are
used as list notation, to distinguish them from the square brackets of
the AVMs.  Named disjunctions are noted as \nd{name}{disj_1 \dis
  disj_2 \dis \ldots\ \dis disj_n}.  A named disjunctions is
interpreted in the following way: In every disjunction with the same
name, the disjunct in the same position has to be chosen.

In the figure~\ref{HNM} the named disjunction {\it v\/} encodes the
described dependence. If the first disjunct of the named disjunction
{\it v\/} is chosen, the finite verb-form is selected, the SUBJ
attribute is assigned the empty list, and the subject as first element
on the COMPS list receives nominative case. In case the second
disjunct is chosen, the VFORM is {\it bse\/}, the SUBJ attribute
contains the singleton list with the subject as element, and only the
complements are encoded in COMPS.  The subject and the complements of
the finite and the non-finite choice are related, since the SUBCAT
list is outside of the disjunction.

If the specification in figure~\ref{HNM} is included in the grammar as
constraint on signs with verbal head value, an example lexical entry
for a verb can be specified as in figure~\ref{HNMentry}.

\myfig{HNMentry}{An example entry for ``love''}{\avmt{word}{PHON &
\nd{w}{\nd{n}{\nd{p}{liebe \dis liebst \dis liebt} \dis \nd{p}{lieben
\dis liebt \dis lieben}} \dis lieben} \\ \path{VFORM} & \nd{w}{\fin
\dis \typ{bse}} \\ \path{SUBCAT} & $<$ \avm{\path{IND} & \idx{1}
\und \avm{NUM & \nd{n}{sg \dis pl} \\ PER & \nd{p}{1st \dis 2nd
\dis 3rd}}} $\;,\;$ \avm{\path{IND} & \idx{2}} $>$ \\ \path{CONT} &
\avmt{love}{LOVER & \idx{1} \\ LOVED & \idx{2}}}}

The lexical entry specifies the semantics and identifies the thematic
roles with the syntactic indices in the standard fashion.
Additionally the named disjunction mechanism takes care of two things.
The disjunction {\it w\/} takes care of the choice between finite and
base phonologies by relating the phonology value to the value of
VFORM. In the case of a finite verb there is an additional choice
regarding the agreement between the index of the subject and the
phonology of the verb. The named disjunction {\it n\/} mediates the
choice between singular and plural, and {\it p\/} links the person
specification on the index to the correct phonology. If instead we are
dealing with a base verb, the named disjunctions {\it n\/} and {\it
  p\/} only occur once, namely in the specification of the subject
index, since the second disjunct of {\it w\/} is chosen. Such a single
occurrence of a named disjunction is equivalent to a single
disjunction. The index of the subject can therefore have any of the
values noted.

The encoding using named disjunctions bears a close resemblance to
encodings on the relational level discussed above, e.g.\ in the form
of definite clause attachments to lexical entries. Nonetheless, there
is a difference worth noting: The named disjunction mechanism is
weaker than the definite clause mechanism in that it does not allow
recursive definitions. Only simple covariation can be expressed, which
is exactly what is needed for the desired encoding.

Summing up, the advantages of the named disjunction mechanism are its
declarative semantics, its restricted expressive power corresponding
to the needs, and the resulting compact encoding of linguistic
generalizations in a computationally usable way. Which class of
lexical generalizations can be captured in this way, should therefore
be interesting to investigate.\footnote{The reader interested in
  constraints on named disjunctions as models for lexical rules should
  note that the example given crucially depends on two things. To
  relate the information of one disjunct of a named disjunction to
  another disjunct of a disjunction with the same name, the disjuncts
  have to structure share with an object outside of the disjunction.
  In the example this is done via the SUBCAT list, which contains the
  subject of both finite and non-finite verbs. The second point to
  mention concerns the disjunction names. These names only make sense
  local to a description. Since the example is based on a constraint
  applying to signs with verbal heads and a description representing
  the lexical entry, the choice of the verb-form has to be repeated in
  the lexical entry itself in order to mediate the choice between the
  constraint and the lexical entry.} For a detailed investigation and
formal definition of named disjunctions, the reader is referred to
\citeN{Eisele&Doerre90} and \citeN{GriffithDiss}.

Concerning the computational systems, it remains to be mentioned
that Troll currently only supports named disjunction of types, while
ALE does not support them at all.

\subsection{Structure licensing} \label{struc-licensing}

In HPSG, constituent structure is licensed via {\it immediate
  dominance schemata\/} and {\it linear precedence rules\/}. Using a
pure constraint logic programming approach to computation, this
mechanism can be directly encoded. Such an approach, however, leaves
us with two problems: one regarding the control structure to be
applied in constraint resolution - since the special guiding status
once attributed to fixed information on phrase structure is now lost
in a vast pool of non-distinguishable constraints; the other in
licensing linear order with LP statements which makes strong demands
regarding the expressive power of the constraint
language.\footnote{Cf.  \citeN{Meurers&Morawietz93} for a discussion
  of the LP formalism.}

The alternative, traditional approach to structure licensing uses the
information about the number of daughters of a certain construction --
and optionally other information specified on constituents together
with the required word order -- and encodes it in {\it phrase
  structure rules\/}. Generalizations over that information are
lost.\footnote{For an approach introducing some word order
  generalizations into a phrase structure grammar, the reader is
  referred to the SUP formalism introduced in
  \citeN{Meurers&Morawietz93}.} The obtained phrase structure
backbone can then be used to drive highly specialized efficient
algorithms to license structures.  Finally, to express some general
constraints on rules, definite clauses can be attached to them.  In a
phrase structure backbone architecture with definite clause
attachments, the number of daughters is the only information which -
compared to an ID/LP approach - has to be additionally
specified.\footnote{The operator ``cats$>$'' supplied by ALE allows to
  specify a list of daughters, which at compile time do not have to be
  of fixed length. Nonetheless it does enforce the number of daughters
  to be fixed at run time.} All the specifications on constituents
distinguishing one constituent from the other and thereby specifying a
certain ordering as well, can theoretically be included in the
definite clause attachments.  Nonetheless parsing algorithms are more
efficient, the more information is present on the constituents.

While the traditional approach has the clear disadvantage of differing
from the ID/LP format used in HPSG theory, it nevertheless leaves the
implementer using the current parser based systems\footnote{In
  principle, parsers can be extended to ID/LP rules and possibly to
  ID/LP schemata as well. Cf.~\citeN{Shieber84},
  \citeN{Seiffert91}, and \citeN{Morawietz94}.} with an interesting
choice regarding generalization vs.\ performance: generalizations over
rules can be captured by encoding the information in a definite clause
which is attached to all the rules generalized over.  The performance
drawback of this encoding is that - since (in general) definite
clauses are executed after a construction has been accepted by the
parser - they do not contribute to the information available to the
parsing algorithm.\footnote{Note that partial execution of definite
  clauses at compile time can make further information available for
  the parser.} This is a disadvantage for processing, since as
mentioned the phrase structure backbone approach is faster the more
information is specified per constituent in a rule, while the exact
algorithm used determines the constituent (e.g.\ leftmost constituent)
on which information specification has the most effect.

\subsubsection{An example for phrase structure rules with the feel of ID
  schemata}
\label{ps-as-id-section}

In the following, one solution to the generalization vs.\ performance
choice is presented using an example from the implementation of HN.
The general idea is to use macros and a definite clause to express the
generalizations over the group of PS rules which replace one ID
schema. Regarding terminology, we will speak of the macro and definite
clause as the {\it core\/} of a {\it group\/} of PS rules which encode
one ID schema.

Generalizations over PS rules are captured in two ways: On the one
hand, the specifications valid for all constituents of a certain
function (e.g.\ head, complement, or mother) are collected in a macro
which is used in all occurrences of a daughter of that function in
every rule of a group.  The specifications specific to a rule are
captured via parameters passed to the macro or by conjoining the
specific descriptions to the call of the macro.  On the other hand,
generalizations over constraints on the whole tree (like the
grammatical principles discussed in the next section) are expressed in
a definite clause which is attached to all PS rules of one group.
These two methods for the specification of phrase structure rules
conserve at least some of the generalizations and intuitions behind
the original ID schemata of the linguistic theory.

To illustrate how the ID/LP constraints of HN were encoded as PS rules
using the method described, let us take a look at the
\typ{head-complement} ID schema.\footnote{In HN verbal projections are
  licensed by the \typ{verbal-complex} and the \typ{head-complement}
  schemata. Because of the valence encoding issue discussed in section~\ref{valence-encoding}, the implementation additionally
  distinguishes a group of \typ{head-subject-complement} rules for
  finite sentences from the \typ{head-complement} group of PS rules,
  which in the implementation only licenses non-finite VPs and finite
  sentences with extracted subjects.} This ID schema licenses
constructions with up to five daughters and a highly varied word order
depending on the specification of the head (e.g.\ AUX and INV
specification on verbal heads) which - as follows from the discussion
above - leads to a high number of phrase structure
rules.\footnote{\setcounter{hilfszaehler}{\value{footnote}}While a
  high number of rules naturally slows down performance, some
  performance can be regained, since some of the rules become so
  specific that it is known which class of lexical heads are possible
  in which rule.  Therefore more information can be specified in the
  rule, speeding up the system.  It should be clear though that this
  additional specification of information already encoded in the
  lexical entries is a theoretically unmotivated repetition. Trying to
  lose this redundancy by only specifying the information in the
  rules, conflicts with the increasing lexicalization going on in
  linguistics today. Phrase structure based formalizations therefore
  seem to be inappropriate for modern lexicalized theories.  Still,
  future pre-compilation techniques might be able to automatically add
  additional specifications to phrase structure rules, which can be
  deduced from interaction with the specification of the lexicon
  used.} The head-complement ID schema of HN is shown in figure
\ref{HN-hc-ID}.\footnote{Because the head-complement ID schema of HN
  depicted in figure~\ref{HN-hc-ID} (= figure 21 of HN) demands the
  occurrence of at least one non-verbal complement, HN also call it
  Head-NP-Complement ID.}

\myfig{HN-hc-ID}{The head-complement ID schema of HN}{
  \avm{SYNSEM\|LOC\|CAT & \avm{HEAD & \typ{verb} \\ COMPS & \elst \\ 
      NPCOMP & plus}} \hspace{0.4em} \idimpl \hspace{1em}
  \parbox[t]{16em}{H \avm{NONLOC\|TO-BIND\|SLASH & \elst},
    \\[1ex] C \avm{SYNSEM\|LOC\|CAT\|HEAD & \(\neg
      verb\)}\(^+\), \\[1ex] (C \avm{SYNSEM\|LOC\|CAT\|HEAD &
      \typ{verb}}).}}

In the implementation the specifications of the mother and the head
category of this ID schema are captured in the following core
definitions, which apply to all PS rules of the head-complement group.

\myfigRef{hc-core}
\begin{verbatim}
hc_mother(S)       macro (@verb,
                          @comps_sat,
                          @plus_npcomp,
                          @val_subj(S), @spr_sat).

hc_head(S,C)       macro (@no_to_slash,
                          @val_subj(S), @spr_sat, @val_comps(C)).\end{verbatim}
\vspace*{-2.6ex}\myfigTitle{The macro core of the \typ{head-comp} group}

Before going through the specifications, a short note on the notation
used might be appreciated.\footnote{The macro mechanism of ALE was
  introduced in section~\ref{lex-classes} on pp.~\pageref{placebo}.
  For a full definition of the ALE syntax, the reader is referred to
  \citeN{ALEmanual}. A set of comments on the additional conventions
  used in our grammar can be found in section~\ref{conventions-section}.} Variable names, serving the same
function as the structure sharing tags in HPSG, are noted with a
capitalized first letter. Names following an ``{\tt @}'' character are
calls to macros, and will be replaced at compile time by the complex
description which the macro abbreviates. The type of the description
abbreviated is noted in the name of the macro. Macros with a suffix
``\_y'' are of type \typ{synsem}, those without suffix (all the macros
in figure~\ref{hc-core}) of type \typ{sign}.  Finally, the commas
within the parenthesis are interpreted as conjunction (except when
separating the arguments of a macro), and the square brackets
represent the normal list notation of PROLOG.

Comparing the core definitions of the implementation with the original
formulation of the Head-Complement ID schema we note a close
similarity: The macro {\tt hc\_mother}, which specifies all the
information on the mother of each PS rule of the head-complement
group, contains the specifications present on the mother of the ID
schema in figure~\ref{HN-hc-ID}. In addition -- for reasons discussed
below -- all valence information is included, which accounts for the
extra {\tt @val\_subj} and {\tt @spr\_sat}. The same holds for the
head: like the head in the ID-schema, the macro {\tt hc\_head}
contains a TO-BIND specification. In addition the valence information
is accessed. The specification of the one optional verbal and the more
than one nominal complements in figure~\ref{HN-hc-ID} cannot be
elegantly expressed in a macro for the complement function. The number
of complements is not fixed in the schema. Still certain properties of
possibly occurring complements need to be specified, which makes it
necessary to fix a certain order of complements. The complements
therefore need to be specified separately in each PS rule of the
head-complement group.

Figure~\ref{ps-example} shows one PS rule of the \typ{head-comp} group
which makes use of the macro core. First, some comments on the ALE
notation used: The operator {\tt ===>} separates the mother of a
phrase structure rule from the daughters. Each daughter is prefixed
with {\tt cat\(>\)} and ends with a comma. Finally, the {\tt
  goal\(>\)} operator allows us to add calls to definite clause
relations to the phrase structure rule defined. The phrase structure
rule can only be used to license a local tree, for which the
relational constraints attached to the rule are satisfiable.

\pagebreak
\myfigRef{ps-example}
\begin{verbatim}
hc_minus_inv2 rule
   (Mother, @hc_mother(S)) ===>
      cat> (C2, @obj_np),
      cat> (C1, @obj_np),
      cat> (Head, @hc_head(S,[C1,C2]), @n_fin, @minus_inv),
      goal> hc_goal(hcni2, Mother, Head, [C1,C2]).\end{verbatim}
\vspace*{-2ex}\myfigTitle{Phrase structure rule of the \typ{head-comp}
  group}

This rule licenses non-inverted non-finite VP constructions in which a
verbal element combines with two complements. For the mother category
of the construction, only the information valid for the whole HC group
(the core specification defined in figure~\ref{hc-core}) is specified
via the call to the macro {\tt @hc\_mother\/}. The variable {\tt
  Mother} is conjoined to that specification to feed the definite
clause {\tt hc\_goal} discussed below. Similarly, the macro {\tt
  @hc\_head\/} is called as description of the head, and the variable
{\tt Head\/} is conjoined to the whole specification to be used in the
call to the definite clause.  Two more things should be mentioned.
For reasons that will be explained in the next section, the valence
principle is compiled into each rule.  The argument list of the mother
macro ({\tt S}) and that of the head macro ({\tt [S],[C1,C2]}) serve
to mediate the valence information.  The conjuncts {\tt @n\_fin} and
{\tt @minus\_inv} in the description of the head daughter specify that
only non-finite non-inverted heads can appear in this rule.  This is
not included in the general {\tt @hc\_head} specification, since the
HC-ID also licenses inverted or finite constructions. The
specification of the complement daughters as nominal complements
results from two sources. The HN specification of the head-complement
ID schema of figure~\ref{HN-hc-ID} demands there to be at least one
non-verbal complement and an inspection of the lexical elements which
can appear as heads in this rule tells us that both complements have
to be object noun-phrases ({\tt @obj\_np}).\footnote{This is a good
  example for the additional specifications in phrase structure rules
  mentioned in footnote \arabic{hilfszaehler}.} At the bottom of the
rule, the goal expressing the constraints applying to all HC
constructions ({\tt @hc\_goal}) is called and the rule specific
specifications (the name of the rule for identification in the parse
tree, the mother, the head, the subject and the list of complements)
are passed as arguments. The definition of the definite-clause core
for the \typ{head-comp} group is shown in figure~\ref{hc-goal}.

\myfigRef{hc-goal}
\begin{verbatim}
hc_goal(Subtype, Mother, Head, Comps) if
   dtrs(Mother, @head_comp_cs(Subtype, Head, Comps)),
   principles(Mother, Head, Head, [Head|Comps], Head).\end{verbatim}
\vspace*{-2ex}\myfigTitle{The definite-clause core of the \typ{head-comp} group}

The call to the {\it dtrs\/} relation serves to build up a parse tree.
It is included in the definite clause core and not in the PS rules in
order to separate it as much as possible from the linguistically
motivated specifications. The call to the principles on the other hand
serves as universal attachment of the grammatical principles. All
information necessary for the principles is supplied via the five
arguments. This allows us to use exactly the same {\it principles\/}
relation for all PS rules. We will see in the next section how the
principles are encoded in this relation.

Finishing up the issue of encoding dominance relations, it remains to
be mentioned that in the grammar the head-complement schema, the
verbal-complex schema, and the extra head-subject-complement schema
are encoded using the mechanism just described. For each of the other
ID schemata in the fragment (marker-head, specifier-head, adjunct-head
and filler-head) there is a one to one correspondence between PS rules
and ID schemata, since they introduce a fixed number of daughters.

\subsection{Expressing the principles}\label{principles}

In the last section we saw two options for specifying constraints on a
construction: directly encoding the constraints in a rule ensuring
maximal performance, and the usage of definite clauses yielding
maximal generality.  When deciding how to encode the principles of a
grammar (e.g.\ the head feature principle or the semantics principle
of \hpsgII) we are again faced with those two
possibilities.\footnote{I here only note the possibilities in a phrase
  structure backbone based system. As discussed in the section~\ref{constraint-section} on expressing constraints, the principles
  in \hpsgII\ are expressed as implicative constraints just like the
  rest of the grammar.  In a general group~1 constraint resolution
  system one has the choice to directly encode the principles as
  object definitions \`a la \hpsgII\ or to compile the principles into
  the object definitions expressing the immediate dominance schemata.
  Choice here is influenced by the strength of constraints that can be
  expressed in the language used (e.g.\ if complex antecedents are
  available) and the question of when which kind of constraint is
  resolved (compile or runtime).}

Due to the decision to stay as close to the linguistic theory as
possible, the head feature principle (HFP), the nonlocal feature
principle (NFP), the semantics principle (SemP), and the marking
principle (MarkP) were each encoded in a separate definite clause and
then conjoined in a clause ``principles'' which then was attached to
all groups of PS rules.  Regarding processing, not specifying these
principles in the phrase structure backbone has only a small negative
effect, since these principles mostly relate information in the mother
to information in the daughters, as opposed to expressing relations
between the daughters.  Since most of the information originates in
the lexicon, the bottom-up strategy employed by the systems used would
only marginally profit from the information mediated by those
principles.  The interface between a rule and the principles consists
of five arguments: Mother, Syntactic\_Head, Semantic\_Head,
List\_of\_Dtrs, and Marking\_Dtr\footnote{The Marking\_Dtr is the
  marker of a head\_marker construction and the head in any other
  construction.}.  This encoding allows an elegant general encoding of
the principles, without having to specify choice procedures (e.g.\ 
picking out the semantic head) depending on the type of the
construction. Figure~\ref{encoded-principles} shows the definition of
the principles in the implemented grammar.

\myfigRef{encoded-principles}
\begin{verbatim}
principles(Mother, Head, Sem_head, Dtr_list, Marking_dtr) if
   (hfp(Mother, Head),
    nfp(Mother, Head, Dtr_list),
    sem_p(Mother, Sem_head, Dtr_list),
    marking_p(Mother,Marking_dtr)).\end{verbatim}
\vspace*{-2ex}\myfigTitle{Definition of the definite clause \typ{principles}}

The head feature principle (HFP) is encoded as displayed in figure
\ref{hfp}.

\myfigRef{hfp}
\begin{verbatim}
hfp(@head(X),@head(X)) if true.\end{verbatim}
\vspace*{-2ex}\myfigTitle{Definition of the definite clause \typ{hfp}}

When we compare this definition with the HFP as given in \hpsgII\ 
(which, for the sake of convenience, is repeated below), we note that
only the consequent of the implication is expressed in the definite
clause. The antecedent is implicit in the encoding, since the definite
clause {\it principles\/} is only attached to PS rules licensing
phrases dominating headed structures.\footnote{The fragment does not
  contain any non-headed structures. Therefore the definite clause
  principles can be attached to all PS rules.}

\myfig{hfp-avm2}{The HFP}{ \avmt{phrase}{DTRS & \hspace{-2ex}
    \typ{headed-struc}} \hspace{1em} \impl \hspace{0.9em}
  \avm{SYNSEM\|LOC\|CAT\|HEAD & \idx{1}
    \\DTRS\|HEAD-DTR\|SYNSEM\|LOC\|CAT\|HEAD & \idx{1}}}

Note that the HFP encoding does not actually make use of the
expressive power of the definite clause mechanism. Were we not
interested in capturing generalizations as discussed above, the HFP
could be encoded directly in the PS rules. This is different for the
nonlocal feature principle (NFP) shown in~\ref{nfp}.

\myfigRef{nfp}
\begin{verbatim}
nfp(@in_slash(Diff), @to_slash(Head_To_slash), Dtr_list) if
   (collect_in_slash(Dtr_list,In_slash),
    list_diff(In_slash,Head_To_slash,Diff)).\end{verbatim}
\vspace*{-2ex}\myfigTitle{Definition of the definite clause {\it nfp\/}}

Again the definition from the appendix of \hpsgII\ is included below.
Since in the fragment we are only concerned with the nonlocal feature
SLASH, the references to the other nonlocal features QUE and REL are
omitted.

\myfig{hpsg2nfp}{The NFP for the nonlocal feature
  SLASH}{\begin{quote} In a phrase whose DAUGHTERS value is of
  sort \typ{headed-structure}, the value of \\
  SYNSEM\|NONLOCAL\|INHERITED\|SLASH is the set difference of the
  union of the values on all the daughters and the value of
  SYNSEM\|NONLOCAL\|TO-BIND\|SLASH on the
  HEAD-DAUGHTER.\end{quote}}

Note that this time it is not easily possible to rewrite this text
definition as an AVM. The ``union of the values on all the daughters''
cannot be directly expressed, since each subtype of
\typ{constituent-structure} has different attributes in which the
daughters of a construction are stored. In the implementation, the
list of all daughters therefore is one of the arguments passed to the
principles, which enables us to use a uniform, recursively defined
relation {\tt collect\_in\_slash} to collect the INHERITED\|SLASH
values of all daughters. Note that -- for reasons explained in section
\ref{slash-as-list-section} -- in the implementation SLASH has a list
instead of a set value. The ``set difference'' of the original
definition therefore becomes the {\tt list\_diff} relation in figure
\ref{nfp}. Without going into the definitions of the two predicates
doing the collecting and the calculation of the difference, it should
be clear that specifying these operations on lists of undetermined
length requires the extra expressive power of the definite clause
mechanism. As a last comment on the encoding of the NFP note that the
call to the macros {\tt @in\_slash} and {\tt @to\_slash} in the first
two arguments of {\tt nfp} serve to select the INHERITED\|SLASH value
of the mother sign, which is passed as the first argument, and the
TO-BIND\|SLASH of the head sign, which is passed as the second.

The valence principle (VP) does not fit well into the schema used for
encoding the principles as definite clause attachments on PS rules.
As mentioned above, the HFP, NFP, SemP, and MarkP mainly relate
information between the mother and the daughters. The VP additionally
relates information between the head and the other daughters. Since
this is valuable information for any processing regime, it is encoded
separately in each group of rules without the usage of definite
clauses. Additional motivation for this dividing up of the valence
principle comes from the fact that it does not easily lend itself to a
general formalization. The formulation in \hpsgII\ uses variables over
attributes and attribute name manipulation (e.g.\ to obtain SUBJ-DTR
from SUBJ) both of which are neither part of the logic behind \hpsgII\ 
as provided by \citeN{King89} or \citeN{Carpenter92} nor of
ALE or Troll.  Encoding the function of a sign in the value of an
attribute of the sign - as proposed in \citeN{Meurers&Morawietz93}
or realized in \citeN{Meurers93} - enables the grammar writer to
refer to the function. All subcategorization requirements can then be
encoded in a single attribute and all daughters of a local tree in a
single list valued daughter attribute of constituent structure without
losing the distinction between the different functions.\footnote{To be
  able to quantify over all daughters - as is necessary to collect the
  nonlocal features in the nonlocal feature principle and the
  quantifiers in the semantics principle - one has to keep track of
  all daughters in a list anyway.  However, as discussed in
  conjunction with the interface specification for the principles,
  this can be done without permanently encoding them using an
  additional attribute.}

\section{Relating the specific linguistic theory to its implementation}
\label{specific-theory}

This section illustrates some issues concerning the specific
linguistic theory of HN which arose in writing the grammar. It
explains where the implementation differs from the original
theory.\footnote{Except for the differences commented on in the
  discussion of examples from the grammar in the previous sections,
  every difference between the linguistic theory and the implemented
  grammar is discussed below. This ensures that the work on the
  implementation of a theory as test for the rigid, explicit, and
  correct formalization of a linguistic theory can provide feedback
  for the further development of the linguistic theory.}

\subsection{Encoding valence}

\label{valence-encoding}

In the discussion on German sentence structure about the existence of
a verb phrase in German, \citeN{PollardHNM} proposes that German
lacks a finite vp while it does have a non-finite one.  This is
technically reflected in the following way\footnote{This is the same
  analysis of finitivization which has already been subject of some
  discussion to illustrate the named disjunction mechanism on pp.\ 
  \pageref{named-disjunction-page}. Only the difference in constituent
  structure involved is relevant in the following discussion.}: The
subject of finite verbs are encoded together with the internal
complements (in attribute SUBCAT), while that of non-finite verbs is
specified in a separate attribute SUBJ.  This lexical information is
put to work via an id schema which realizes all elements in one local
tree which are subcategorized for in SUBCAT.  A single ID schema now
suffices to build finite sentences and verb phrases and - since no ID
schema ever refers to the SUBJ attribute - one explains that only
subjects of finite verbs can be realized. HN adopt this elegant
technique.\footnote{Note a minor notational change. HN use a COMPS
  attribute in the style of chapter 9 of \hpsgII\ instead of the
  SUBCAT attribute used in \citeN{PollardHNM}.}

From an implementor's point of view the proposal loses at least some
of its elegance. As referred to in section~\ref{lexicon}, HN propose
an argument raising specification as part of the lexical entries of
auxiliary verbs.  Figure~\ref{HNarg-raise} shows the basic form of the
argument raising specification of HN.\footnote{HN use a functional
  notation for the append relation in their diagrams.}

\myfig{HNarg-raise}{The HN argument raising specification on a non
  finite auxiliary (\typ{valence} value and relational attachment
  shown)}{\avm{COMPS & \idx{3}} \hspace{1em} \& \hspace{1em}
  append(\idx{1}, \avm{SYNSEM\|LOC\|CAT & \avm{HEAD & \typ{verb} \\ 
      COMPS & \idx{1}}}, \idx{3})}

Looking at figure~\ref{HNarg-raise}, one sees that argument raising
introduces a tag in the lexical entry (\idx{3}) which in the lexicon
is completely unrestricted. The set of objects described by the first
argument and the result argument of the definite clause {\it append\/}
only get narrowed down by the usage of the lexical entry in a tree.
The HN use of append in the lexical entry of auxiliary verbs therefore
presupposes a rather sophisticated control mechanism for constraint
resolution which executes a definite clause at the time its essential
arguments are instantiated. Neither ALE nor Troll offered such control
mechanisms at the time of the implementation.\footnote{The delay
  patterns of CUF are an example for a mechanism which would handle at
  least these kind of cases (cf.  \citeN{Doerre&Dorna93}).  The
  newest version of Troll offers a rather sophisticated freeze
  mechanism as well.}

If the order of the COMPS list is reversed, the usual encoding of
lists (in which the first element and the rest of a list can be
directly accessed) results in an encoding which no longer needs the
{\it append\/} relation. We can simply structure share the embedded
verb's COMPS with the tail of the outer COMPS list.  This change of
the valence list order further reduces complexity by eliminating the
calls to the definite clauses which were needed in local trees to
access the last, most oblique element first since those usually get
realized first.

However, ordering the elements on the subcategorization list by
decreasing obliqueness (as in \citeN{hpsg1}) is also problematic;
again because of the argument raising mechanism.  To see this, the
argument raising specification with inverted comps list (cf.\ figure
\ref{arg-raise}) is repeated as~\ref{arg-raise-copy} below.

\myfig{arg-raise-copy}{Argument raising specification on a non-finite
  auxiliary (\typ{valence} value shown)}{\avm{COMPS &
    \lst{\avm{SYNSEM\|LOC\|CAT & \avm{HEAD & \typ{verb} \\ COMPS &
          \idx{1}}}}{\idx{1}}}}

If we try to formulate a lexical rule forming a finite auxiliary in
the style of \citeN{PollardHNM} on the basis of the valence
specification in~\ref{arg-raise-copy}, we are stuck with the
impossible: trying to add the subject as the least oblique element to
the end of a list of undetermined length. Even the desperate move of
putting the subject into the head of the list (which corrupts the
theoretically motivated obliqueness ordering) is no solution, since
then - due to the now in general undetermined position of a possibly
occurring verbal-complement - we need to separate phrase structure
rules for finite and non-finite verbs, or for auxiliaries and main
verbs. This, however, is exactly which we intended to avoid via
encoding the subject on COMPS.

Concerning the implementation and in the light of the problems with
an argument raising specification using a call to append, it therefore
seems to be cleaner to keep the subject encoded in SUBJ for non-finite
and finite verbs, which allows easy list manipulation with a COMPS
list ordered by decreasing obliqueness.

Since under this approach one needs to separately license
constructions that realize subjects and those that do not, a {\it
  head-subject-complement\/} ID schema is introduced in the grammar to
license finite sentences containing subjects. The
\typ{head-complement} ID of the implemented grammar therefore only
licenses non-finite VPs and finite sentences with extracted subjects.

\subsection{Specifying SLASH sets depending on COMPS lists}
\label{slash-as-list-section}

A problem similar to the use of append in the lexical entries of
auxiliaries discussed above arises from the formulation of the
PVP-Topicalization Lexical Rule in HN. In order to relate the
list-valued COMPS list to a set-valued SLASH list HN attach a relation
\typ{same-members} to the lexical rule (cf.\ figure 19 of HN). The
definite clause \typ{same-members} can only be executed once it is
clear which constituents have been syntactically realized where. In
the terminology of lexical rules of meta-relations, the same-members
specification therefore only works with unification as application
criterion. Regarding the computational treatment of lexical rules, it
demands that the lexical rules be not completely compiled out, but at
least partly computed ``on demand'' while building the syntactic tree.
Even if this should turn out to be possible\footnote{The discussion of
  the lexical rule mechanism in section~\ref{lex-rules} suggests that
  there are some severe theoretical and computational problems for
  such an approach.} such an approach would demand a highly
sophisticated control structure.

Since one might want to argue anyway that the embedding constraint on
extraction is to be captured in syntax, making the value of SLASH a
last-in-first-out list store is a possible solution to our problem,
since it allows us to replace the \typ{same-members} relation in the
PVP extraction lexical rule by a simple token-identity constraint.
The pragmatic solution chosen for the grammar implemented therefore is
to change the value of the SLASH attribute to be a list instead of a
set. The resulting lexical rule was already shown in figure
\ref{hn-pvp} of p.~\pageref{hn-pvp}.

\subsection{The Cooper-storage mechanism}

In the \hpsgII\ theory a Cooper-storage mechanism is used to calculate
all scopings of the quantifiers in a sentence. Since this mechanism is
integrated into the HPSG setup just like the rest of the constraints
making up the grammar, in a straightforward implementation all
scopings of quantifiers will be computed at the same time as the
grammatical constraints are applied to a local tree. This leads to the
same local tree being admitted several times, the only difference
between the trees being the number of quantifiers retrieved. This is
an undesirable effect, since these local trees are identical for all
syntactic properties that play a role regarding the grammaticality of
the trees. 

The concept of an underspecified discourse representation defined in
\citeN{Reyle93} replaces the forced computation of all possible
quantifier scopes with a representation of quantifiers underspecified
for scope. Restrictions on the scope relations can then be formulated
on the basis of the underspecified representation. This setup avoids
an unmotivated multiplication of syntactic structures while at the
same time introducing an elegant mechanism for expressing theories
about the semantic scope of quantifiers.  \citeN{Frank&Reyle92} show
that this setup can be transferred to HPSG and implemented in an HPSG
grammar.

Since the implementation of HN documented here is concerned with
syntactic phenomena only, the semantics included in the grammar simply
follows the proposals made in \hpsgII. However, in the semantics
principle an additional specification ensures that no quantifiers are
ever retrieved.  Every sentence therefore contains the collection of
all its quantifiers in its QSTORE attribute as a simple kind of a
representation underspecified for quantifier scope. This way we obtain
a workable setup avoiding the multiple structures with the help of a
minor modification.\footnote{Naturally, the modification of the
  semantics principle made is not intended to be a theoretical
  proposal. It merely is a loose link between an \hpsgII-style
  semantics and an underspecified representation of quantifiers
  inspired from \citeN{Reyle93} and \citeN{Frank&Reyle92}. In any
  more semantic-oriented grammar fragment that link would have to be
  firmly established.}

\subsection{Technical comments on the implementation}

The appendix contains the complete ALE grammar including a collection
of test sentences for all relevant constructions. Additionally the
printout of a test run is included to illustrate the performance of
the grammar in the ALE system.

\subsubsection{The lexicon}

The lexicon of the implementation contains the entries necessary to
follow the analyses of aux-flip and partial verb phrase topicalization
in V1, V2, and VL sentences. As discussed in section
\ref{lex-classes}, the implementation contains a hierarchy of lexical
macros allowing an easy extension of the lexicon. All lexical entries
are specified using calls to the lexical macros.

After lexical rule application, the lexicon contains the following
entries:

\mysection{Complementizer} {\it da{\ss} (that)}

\mysection{Adjectives} {\it klein (small), gut (good)}

\mysection{Determiners} definite: {\it der (the)}; indefinite: {\it
ein (a), einige (some)};  negative indefinite: {\it kein (no)}

\mysection{Nouns} {\it Mann (man), Tisch (table), Buch (book),
  Geschichte (story), Verwandter (relative), Karl, Peter, Anna, Sarah}

Adjectives, determiners and nouns are included in all possible
number, gender, case, and declension pattern specifications.

\mysection{Main verbs} intrans: {\it lachen (laugh)}; trans: {\it
  lieben (love),
finden (find), lesen (read)}; ditrans: {\it geben (give), erz\"ahlen (tell)}

\mysection{Auxiliaries} {\it haben (have), werden (will), k\"onnen
  (can), m\"ussen (must), wollen (want to)}

Verbs are given in their third person singular and plural, finite,
past participle\footnote{Only past participle formation with ``haben''
  (have) is included. Therefore the past participle of {\it werden\/}
  is not part of the lexicon.}, base forms, and (for auxiliaries) the
substitute infinitive forms.

\subsubsection{Macros and notational conventions}
\label{conventions-section}

Many abbreviations used by HPSG linguists serve to provide easy access
to information deeply embedded in descriptions.  The grammar contains
an abbreviation of type \typ{sign} and one of type \typ{synsem} (which
are the types of objects most often talked about in the grammar) for
all relevant properties.  The result is that the descriptions in the
implementation often closely resemble the specifications in HN.

Since macros contain descriptions of objects of a specific type, the
type of a macro is encoded in its name.  Suffix ``\_y'' is used for
macros of type \typ{synsem} and macros without suffix are of type
\typ{sign}.  Those macro names ending in ``\_lex'' are macros of type
\typ{sign} belonging to the lexical hierarchy used to specify lexical
entries.

It would be of great value to the user if systems would automatically
infer the correct feature path at which a specification
belongs.\footnote{A similar suggestion was made by Jo Calder at the
  1st International HPSG Workshop in Columbus, Ohio, July/August 93.}
For example the description a) below could automatically be expanded
to that given in b).

\begin{itemize}
\item [a)] \(word \wedge verb \wedge plus\_aux\)
\item [b)] \avmt{word}{SYNSEM\|LOC\|CAT\|HEAD & \avmt{verb}{AUX & plus}}
\end{itemize}

Clearly some notational conventions (e.g.\ for the \typ{plus\_aux}
case above) and some decisions for the search strategy (e.g.\ shortest
path wins) have to be made, but there seem to be no problems which
render such an extension impossible. Such a notational mechanism would
eliminate many of the usages of macros and the syntactic errors
commonly made by the grammar writer in defining those macros.


\newpage
\label{references}

\newpage
\thispagestyle{empty}
\section*{Appendix A: The grammar}
(The appendices are in the file ``appendix.tex''.)
\label{appendix-grammar}

\parindent=2em
\begin{thebibliography}{}

\bibitem[\protect\citeauthoryear{A\"{\i}t-Kaci}{A\"{\i}t-Kaci}{1984}]{Ait-Kaci%
84}
A\"{\i}t-Kaci, H. (1984).
\newblock {\em A lattice theoretic approach to computation based on a calculus
  of partially ordered type structures}.
\newblock Ph.\ D. thesis, University of Pennsylvania.

\bibitem[\protect\citeauthoryear{Carpenter}{Carpenter}{1992}]{Carpenter92}
Carpenter, B. (1992).
\newblock {\em The logic of typed feature structures}, Volume~32 of {\em
  Cambridge Tracts in Theoretical Computer Science}.
\newblock Cambridge University Press.

\bibitem[\protect\citeauthoryear{Carpenter}{Carpenter}{1993}]{ALEmanual}
Carpenter, B. (1993, May).
\newblock {ALE} -- The Attribute Logic Engine, User's Guide.
\newblock Laboratory for Computational Linguistics, Philosophy Department,
  Carnegie Mellon University, Pittsburgh, {PA} 15213.

\bibitem[\protect\citeauthoryear{D\"orre and Dorna}{D\"orre and
  Dorna}{1993}]{Doerre&Dorna93}
D\"orre, J. and M.~Dorna (1993, August).
\newblock {CUF} -- a formalism for linguistic knowledge representation.
\newblock In J.~D\"orre (Ed.), {\em Computational aspects of constraint based
  linguistic descriptions I}, pp.\  1--22. Universit\"at Stuttgart: DYANA-2
  Deliverable R1.2.A.

\bibitem[\protect\citeauthoryear{Eisele and D{\"{o}}rre}{Eisele and
  D{\"{o}}rre}{1990}]{Eisele&Doerre90}
Eisele, A. and J.~D{\"{o}}rre (1990).
\newblock Disjunctive Unification.
\newblock Technical Report 124, {IBM Wissenschaftliches Zentrum, Institut
  f\"{u}r Wissensbasierte Systeme}.

\bibitem[\protect\citeauthoryear{Frank and Reyle}{Frank and
  Reyle}{1992}]{Frank&Reyle92}
Frank, A. and U.~Reyle (1992).
\newblock How to Cope with Scrambling and Scope.
\newblock In G.~G\"orz (Ed.), {\em Konvens 92, 1.\ Konferenz ``Verarbeitung
  nat\"urlicher Sprache''}, Berlin. Springer.

\bibitem[\protect\citeauthoryear{Gazdar, Klein, Pullum, and Sag}{Gazdar
  et~al.}{1985}]{gkps}
Gazdar, G., E.~Klein, G.~Pullum, and I.~Sag (1985).
\newblock {\em Generalized phrase structure grammar}.
\newblock Cambridge, Massachusetts: Harvard University Press.

\bibitem[\protect\citeauthoryear{Gerdemann and King}{Gerdemann and
  King}{1993}]{Gerdemann&King93}
Gerdemann, D. and P.~J. King (1993).
\newblock Typed Feature Structures for Expressing and Computationally
  Implementing Feature Cooccurrence Restrictions.
\newblock In {\em Proceedings of 4. Fachtagung der Sektion Computerlinguistik
  der Deutschen Gesellschaft {f\"ur} Sprachwissenschaft}, pp.\  33--39.

\bibitem[\protect\citeauthoryear{Gerdemann and King}{Gerdemann and
  King}{1994}]{Gerdemann&King94}
Gerdemann, D. and P.~J. King (1994).
\newblock The Correct and Efficient Implementation of Appropriateness
  Specifications for Typed Feature Structures.
\newblock In {\em Proceedings of {COLING-94}. Kyoto, Japan}.

\bibitem[\protect\citeauthoryear{G\"otz}{G\"otz}{1994}]{Goetz94}
G\"otz, T.~W. (1994).
\newblock A normal form for typed feature structures.
\newblock Arbeitspapiere des {SFB} 340 Nr.\ 40, Universit\"at T\"ubingen.

\bibitem[\protect\citeauthoryear{Griffith}{Griffith}{ming}]{GriffithDiss}
Griffith, J. (forthcoming).
\newblock {\em Disjunction and efficient processing of feature descriptions}.
\newblock Ph.\ D. thesis, Universit\"at T\"ubingen.
\newblock Tentative Title.

\bibitem[\protect\citeauthoryear{Hinrichs and Nakazawa}{Hinrichs and
  Nakazawa}{1991}]{Hinrichs&Nakazawa91}
Hinrichs, E.~W. and T.~Nakazawa (1991).
\newblock Linearizing Finite {AUX} in {German} Complex {VP}s.
\newblock To appear in: Nerbonne, John, Klaus Netter and Carl Pollard (eds.):
  {\it {German} Grammar in {HPSG}}, CSLI Lecture Notes.
\newblock Paper presented at the Conference '{HPSG} in {German}' held at the
  University of the Saarland, Saarbr\"ucken.

\bibitem[\protect\citeauthoryear{Hinrichs and Nakazawa}{Hinrichs and
  Nakazawa}{1994}]{Hinrichs&Nakazawa94}
Hinrichs, E.~W. and T.~Nakazawa (1994).
\newblock Partial-{VP} and {S}plit-{NP} Topicalization in {G}erman - An {HPSG}
  Analysis.
\newblock In: Erhard W.\ Hinrichs, W.\ Detmar Meurers, and Tsuneko Nakazawa:
  {\it Partial-{VP} and Split-{NP} Topicalization in {German} -- An {HPSG}
  Analysis and its Implementation}. Arbeitspapiere des {SFB} 340 Nr.\ 58,
  Universit\"at T{\"u}bingen.

\bibitem[\protect\citeauthoryear{H\"ohfeld and Smolka}{H\"ohfeld and
  Smolka}{1988}]{Hoehfeld&Smolka88}
H\"ohfeld, M. and G.~Smolka (1988).
\newblock Definite relations over constraint languages.
\newblock {LILOG} technical report, number 53, {IBM} Deutschland {GmbH}.

\bibitem[\protect\citeauthoryear{H\"ohle}{H\"ohle}{1994}]{Hoehle94}
H\"ohle, T.~N. (1994).
\newblock Spurenlose {Extraktion}.
\newblock Handout of June 1, 1994, Universit\"at T\"ubingen.

\bibitem[\protect\citeauthoryear{Jaffar and Lassez}{Jaffar and
  Lassez}{1987}]{Jaffar&Lassez87}
Jaffar, J. and J.-L. Lassez (1987).
\newblock {Constraint Logic Programming}.
\newblock In ACM (Ed.), {\em Proceedings of the 14th ACM Symposium on
  Principles of Programming Languages, Munich}, pp.\  111--119.

\bibitem[\protect\citeauthoryear{Johnson}{Johnson}{1988}]{Johnson88}
Johnson, M. (1988).
\newblock {\em Attribute-value Logic and the Theory of Grammar}, Volume~16 of
  {\em CSLI Lecture Notes}.
\newblock Stanford, CA: CSLI.

\bibitem[\protect\citeauthoryear{Keller}{Keller}{1993}]{Keller93}
Keller, F. (1993).
\newblock Encoding {HPSG} Grammars in {TFS}, Part {III} - Encoding Revised
  {HPSG}.
\newblock ms., Universit\"at Stuttgart.

\bibitem[\protect\citeauthoryear{King}{King}{1989}]{King89}
King, P.~J. (1989).
\newblock {\em A logical formalism for head-driven phrase structure grammar}.
\newblock Ph.\ D. thesis, University of Manchester.

\bibitem[\protect\citeauthoryear{King}{King}{1994}]{King94b}
King, P.~J. (1994).
\newblock An Expanded Logical Formalism for Head-driven Phrase Structure
  Grammar.
\newblock Arbeitspapiere des {SFB} 340, Universit\"at T\"ubingen.

\bibitem[\protect\citeauthoryear{King and G\"otz}{King and
  G\"otz}{1993}]{King&Goetz93}
King, P.~J. and T.~W. G\"otz (1993).
\newblock Eliminating the feature introduction condition by modifying type
  inference.
\newblock Arbeitspapiere des {SFB} 340 Nr.\ 31, Universit\"at T\"ubingen.

\bibitem[\protect\citeauthoryear{Kuhn}{Kuhn}{1993}]{Kuhn93}
Kuhn, J. (1993).
\newblock Encoding {HPSG} Grammars in {TFS}, {Part} {I} \& {II} - {Tutorial}.
\newblock ms., Universit\"at Stuttgart.

\bibitem[\protect\citeauthoryear{Manandhar}{Manandhar}{1993}]{Manandhar93}
Manandhar, S. (1993, August).
\newblock {CUF} in Context.
\newblock In J.~D\"orre (Ed.), {\em Computational Aspects of Constraint Based
  Linguistic Descriptions I}, pp.\  43--56. Universit\"at Stuttgart: DYANA-2
  Deliverable R1.2.A.

\bibitem[\protect\citeauthoryear{Meurers}{Meurers}{1993}]{Meurers93}
Meurers, W.~D. (1993).
\newblock A {TFS} Implementation of {C}. {P}ollard: ``{O}n {H}ead
  {N}on-{M}ovement''.
\newblock commented implementation. SFB 340/B4, Universit\"at T\"ubingen.

\bibitem[\protect\citeauthoryear{Meurers and G\"{o}tz}{Meurers and
  G\"{o}tz}{1993}]{Meurers&Goetz93}
Meurers, W.~D. and T.~G\"{o}tz (1993).
\newblock Using {TFS} and {CUF}.
\newblock Handout of May 10, 1993, SFB 340/B4, Universit\"at T\"ubingen.

\bibitem[\protect\citeauthoryear{Meurers and Morawietz}{Meurers and
  Morawietz}{1993}]{Meurers&Morawietz93}
Meurers, W.~D. and F.~Morawietz (1993).
\newblock An Alternative Approach to {ID/LP} - {SUP} Grammars.
\newblock Paper presented at the 1st International Workshop on Head-Driven
  Phrase Structure Grammar held at Columbus, Ohio.

\bibitem[\protect\citeauthoryear{Morawietz}{Morawietz}{1994}]{Morawietz94}
Morawietz, F. (1994).
\newblock Direct Parsing of (Typed) Unification ID/LP Grammars.
\newblock Master's thesis, Seminar f\"ur Sprachwissenschaft, Universit\"at
  T\"ubingen, T\"ubingen.

\bibitem[\protect\citeauthoryear{Moshier and Rounds}{Moshier and
  Rounds}{1987}]{Moshier&Rounds87}
Moshier, M.~A. and W.~C. Rounds (1987).
\newblock A Logic for Partially Specified Data Structures.
\newblock In {\em Proceedings of the 14th {ACM} Symposium on Principles of
  Programming Languages. M\"unchen, Germany}, pp.\  156--167.

\bibitem[\protect\citeauthoryear{Pollard}{Pollard}{1990}]{PollardHNM}
Pollard, C. (1990).
\newblock On Head Non-Movement.
\newblock In {\em Proceedings of the Tilburg Conference on Discontinuous
  Constituents}.
\newblock also in: Horck, Arthur and Wietske Sijtsma (in press): "Discontinuous
  Constituency"; Berlin: Mouton de Gruyter.

\bibitem[\protect\citeauthoryear{Pollard}{Pollard}{1993}]{Pollard93}
Pollard, C. (1993).
\newblock Lexical rules and metadescriptions.
\newblock Handout of October 5, 1993, Universit\"at Stuttgart.

\bibitem[\protect\citeauthoryear{Pollard and Sag}{Pollard and
  Sag}{1987}]{hpsg1}
Pollard, C. and I.~Sag (1987).
\newblock {\em Information-Based Syntax and Semantics, Vol. 1: Fundamentals}.
\newblock CSLI Lecture Notes 13. Stanford, CA: CSLI.

\bibitem[\protect\citeauthoryear{Pollard and Sag}{Pollard and
  Sag}{1994}]{hpsg2}
Pollard, C. and I.~Sag (1994).
\newblock {\em Head-Driven Phrase Structure Grammar}.
\newblock Chicago: University of Chicago Press and Stanford: CSLI Publications.
\newblock (version of Feb 1, 1994).

\bibitem[\protect\citeauthoryear{Reyle}{Reyle}{1993}]{Reyle93}
Reyle, U. (1993).
\newblock Dealing with Ambiguities by Underspecification: Construction,
  Representation and Deduction.
\newblock {\em Journal of Semantics\/}, 123--179.

\bibitem[\protect\citeauthoryear{Rounds and Kasper}{Rounds and
  Kasper}{1986}]{Rounds&Kasper86}
Rounds, W.~C. and R.~T. Kasper (1986).
\newblock A Complete Logical Calculus for Record Structures Representing
  Linguistic Information.
\newblock In {\em Proceedings of the 15th Annual {IEEE} Symposium on Logic in
  Computer Science, Cambridge, {MA}, {USA}}, pp.\  38--43.

\bibitem[\protect\citeauthoryear{Seiffert}{Seiffert}{1991}]{Seiffert91}
Seiffert, R. (1991).
\newblock Unification--ID/LP Grammars: Formalization and Parsing.
\newblock In O.~Herzog and C.-R. Rollinger (Eds.), {\em Text Understanding in
  LILOG}, pp.\  63--73. Berlin: Springer.

\bibitem[\protect\citeauthoryear{Shieber}{Shieber}{1984}]{Shieber84}
Shieber, S.~M. (1984).
\newblock Direct Parsing of {ID}/{LP} Grammars.
\newblock {\em Linguistics and Philosophy\/}~{\em 7}, 135--154.

\bibitem[\protect\citeauthoryear{Shieber}{Shieber}{1986}]{Shieber86}
Shieber, S.~M. (1986).
\newblock {\em An Introduction to Unification-Based Approaches to Grammar}.
\newblock CSLI Lecture Notes No. 4. Chicago: Chicago University Press.

\bibitem[\protect\citeauthoryear{Smolka}{Smolka}{1988}]{Smolka88}
Smolka, G. (1988).
\newblock A Feature Logic with Subsorts.
\newblock {LILOG} technical report, number 33, {IBM} Deutschland {GmbH}.

\end{thebibliography}
\end{document}


\pagenumbering{roman}
\section*{Appendix A: The grammar}
\begin{footnotesize}

\newpage
\section*{Appendix B: A test run}
For this test run, ALE 1.0 (release 1 May 1993) under SICStus 2.1 \#9
was used on a Sun SPARC 10 with two 40MHz Super Sparc CPUs and 48Mb
Ram.
%
\end{footnotesize}